\pdfoutput=1
\documentclass[11pt]{article}
\usepackage[letterpaper,top=1in,bottom=1.1in,left=1.05in,right=1.05in]{geometry}
\usepackage{times}
\usepackage{xcolor}
\usepackage[numbers,sort&compress]{natbib}
\usepackage{titlesec}
\usepackage[font=small,labelfont=bf]{caption}

\usepackage{amsmath,amsfonts,bm}

\def\eqref#1{equation~\ref{#1}}

\def\1{\bm{1}}

\DeclareMathAlphabet{\mathsfit}{\encodingdefault}{\sfdefault}{m}{sl}
\SetMathAlphabet{\mathsfit}{bold}{\encodingdefault}{\sfdefault}{bx}{n}

\DeclareMathOperator*{\argmax}{arg\,max}

\usepackage{hyperref}
\definecolor{accent}{RGB}{31,90,166}
\hypersetup{colorlinks=true, linkcolor=accent, citecolor=accent, urlcolor=accent}
\titleformat{\section}{\large\bfseries}{\thesection.}{0.6em}{}
\titlespacing*{\section}{0pt}{16pt plus 3pt minus 2pt}{8pt plus 1pt}
\titleformat{\subsection}{\normalsize\bfseries}{\thesubsection.}{0.55em}{}
\titlespacing*{\subsection}{0pt}{12pt plus 2pt minus 2pt}{6pt plus 1pt}
\titleformat{\paragraph}[runin]{\normalsize\bfseries}{}{0pt}{}[.]
\titlespacing*{\paragraph}{0pt}{9pt plus 2pt minus 1pt}{0.6em}
\usepackage{url}
\usepackage{graphicx}
\usepackage{booktabs}
\usepackage{makecell}
\usepackage{multirow}
\usepackage{tabularx}
\usepackage{amsmath,amssymb}
\usepackage{amsthm}
\usepackage{colortbl}
\usepackage{float}
\usepackage{tikz}
\usepackage{algorithm}
\usepackage{algorithmic}
\usepackage{fvextra}
\usepackage{pifont}
\usepackage[most]{tcolorbox}
\usetikzlibrary{arrows.meta,positioning}

\definecolor{wingreen}{RGB}{0,128,60}
\definecolor{tickgreen}{RGB}{46,139,87}
\definecolor{softcross}{RGB}{205,92,92}
\providecommand{\yes}{\textcolor{tickgreen}{\ding{51}}}
\providecommand{\no}{\textcolor{softcross}{\ding{55}}}
\newcommand{\gain}[1]{\textcolor{wingreen}{\textbf{+#1}}}
\newcommand{\method}{Self-Reflection over Executable Reasoning}
\newcommand{\methodshort}{executable reflection}
\newcommand{\cwi}{Code-with-Image}
\newcommand{\bench}{CwI-Bench}

\newcommand\blfootnote[1]{\begingroup\renewcommand\thefootnote{}\footnote{#1}\addtocounter{footnote}{-1}\endgroup}  
\newtcolorbox{promptbox}[1]{enhanced, breakable, colback=black!4, colframe=black!70,
  boxrule=0.5pt, arc=2mm, left=3mm, right=3mm, top=2mm, bottom=2mm,
  title={#1}, fonttitle=\bfseries, coltitle=white, colbacktitle=black!70,
  attach boxed title to top left={yshift=-2mm, xshift=3mm}, boxed title style={arc=1mm}}

\newtcolorbox{abstractbox}{enhanced, colback=accent!6, colframe=accent!6, boxrule=0pt,
  arc=2.5mm, left=5mm, right=5mm, top=4mm, bottom=4mm}

\begin{document}
\noindent{\LARGE\bfseries Self-Evolving \textcolor{accent}{Code-with-Image} Reasoning\par}
\vspace{10pt}
\noindent{\normalsize Tianze Yang\textsuperscript{1,2}, Liang Wu\textsuperscript{1}, Ruitong Sun\textsuperscript{2}, Yucheng Shi\textsuperscript{2}, Yanqiao Wang\textsuperscript{2}, Mayank Darbari\textsuperscript{1}, Ninghao Liu\textsuperscript{3}, Jin Sun\textsuperscript{2}, Liangjie Hong\textsuperscript{1}\par}
\vspace{4pt}
\noindent{\small \textsuperscript{1}Nokia\quad\textsuperscript{2}University of Georgia\quad\textsuperscript{3}The Hong Kong Polytechnic University\par}
\vspace{3pt}
\noindent{\small \raisebox{-2.2pt}{\includegraphics[height=11pt]{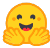}}\hspace{4pt}\href{https://huggingface.co/datasets/ytz009/image_pot_bench}{Hugging Face: CwI-Bench}\hspace{14pt}\raisebox{-2.2pt}{\includegraphics[height=11pt]{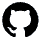}}\hspace{4pt}\href{https://code-with-image.github.io}{Project Website}\par}
\blfootnote{Work conducted during an internship at Nokia, Sunnyvale, CA, USA.}
\blfootnote{Correspondence: \texttt{tianze.yang@nokia.com}}
\vspace{6pt}

\begin{abstractbox}\small
Multimodal models increasingly reach for tools when solving visual tasks (crop, zoom, rotate, brighten), a paradigm known as \emph{thinking-with-images}. The central challenge is one of perception: tools mostly serve to expose visual evidence, reasoning over that evidence stays in language, and most targets are ones a human could in principle determine by inspection. Some visual questions, however, are not bottlenecked by perception: recovering their answers requires executing a multi-step visual algorithm over the pixels. On such questions a model often names the correct algorithm at once yet still answers wrong, because language can describe an algorithm without being able to run one. \emph{\cwi{}} crosses that line: given nothing but a Python interpreter, the model must implement a genuine visual algorithm in code to solve the task; the program itself becomes the reasoning. The bottleneck then shifts from executing code to deciding which algorithm to implement. So we let the model teach itself: a training-free reflection loop studies its own failed programs, tests repairs against constructive ground truth, and keeps what survives as portable skills. On our \emph{Code-with-Image Bench} (\bench{}), thirty task families induced by hidden visual computations with disjoint learning and evaluation splits, even GPT-5.6-luna stays below $30\%$ with tool-free chain of thought; given a bare interpreter it reaches $43\%$, and with skills evolved through its own executable reflection, $67\%$. The open 27B model climbs the same ladder ($9\%\!\to\!33\%\!\to\!56\%$), and the skills are plain text, transferable across scales and families. When code carries the reasoning, debugging code becomes debugging reasoning.
\end{abstractbox}

\section{Introduction}
\label{sec:intro}
Multimodal models are increasingly asked visual questions that no single glance can settle. The \emph{thinking-with-images} paradigm \citep{twisurvey,deepeyes,deepeyesv2,pixelreasoner,pyvision,thyme,codev,codedance} responds by letting the model act: instead of answering in one pass, it iteratively manipulates visual evidence and invokes tools across many steps. It crops and zooms until a small target becomes legible \citep{vstar,deepeyes,pixelreasoner}, sketches marks as visual thoughts \citep{hu2024sketchpad}, applies image preprocessing such as rotation, brightness, or contrast adjustment \citep{thyme,pyvision}, or pairs image operations with web search when the pixels alone cannot answer \citep{deepeyesv2,agenticmme}. Whatever the action, its role is the same: tool use fills single-step gaps in a linguistic chain (a sharper view where the eye fails, a fact where knowledge fails, a value where arithmetic fails), and reasoning resumes in language over what the tool returns. The ceiling of this division is becoming visible: $93$--$96\%$ of the problems SOTA agents of this line solve with tools are also solved without them \citep{tooluseaudit}, and tool use adds only $2$--$5\%$ accuracy on the very suites built to host it \citep{codevision}.

That ceiling has a reason: some visual questions cannot be answered by any chain of sentences, however good the evidence handed back. Consider image forensics: has this photograph been tampered with, and how? Were two exposures blended? Were the color channels remixed? By how many pixels is a spliced region misaligned? Such questions have exact answers, fully determined by the pixels, yet looking does not reveal them. Recovering one takes a chain of computations over the whole image, and the intermediate results are arrays, not sentences: each step consumes the previous step's output in full. A strong model can usually name the right procedure; it cannot run it, because a chain of thought moves words while the procedure moves data. Familiar visual tools do not help either: there is nothing here to crop, zoom into, or search for. Two things are missing. The first is a medium: reasoning must move from language into something that can hold and transform arrays. The second is the algorithm itself: each question needs its own, no fixed toolbox can be prepared in advance, and so the model has to write it.

\emph{\cwi{}} moves the reasoning to where its steps live. The environment supplies exactly one thing, a general-purpose Python interpreter; the image enters it as data, the chain of transformations is written as a program, and executing the program is performing the reasoning: coding is the reasoning process. In thinking with images, tools produce evidence and language produces the answer; in \cwi{}, language produces a program and the program produces the answer. Agents that write code already exist \citep{pyvision,thyme,codev,codedance}; what has been missing is not the capability but the focus: making the program-carried solution itself the object of study.

To isolate this regime we build \bench{}, the Code-with-Image Bench: $30$ task families, each induced by a latent visual computation (shifts, permutations, localization, color mixing, signal measurements) whose generator labels instances constructively and without bound, with disjoint train, validation, and test splits. Every target lies \emph{beyond inspection}: at the scored precision no amount of looking, zooming, or searching determines the answer. No reference toolchain exists to imitate, so tool-free failure isolates missing computation, and the suite measures whether an agent improves at a visual computation, not merely whether it performs one. In language alone, GPT-5.6-luna stays below $30\%$ even in thinking mode.

Access to an executable substrate is not enough: execution stops being the bottleneck, and deciding what code to write becomes it. The model still fails on silent implementation faults (unsigned arithmetic wraps, a seam index is off by one), none of which its own prose betrays. These failures motivate the question: can the model find the causes of its own failures, repair them, and carry the repair to similar problems without weight updates or human-written procedures?

Our answer is \method{}, a loop with two modes. \emph{Observational reflection} reads multimodal trajectories and proposes bounded edits to a skill library; when reading stops helping, \emph{executable reflection} reloads failed images and code into a sandbox, runs diagnostics, renders intermediates, and tests repairs. What survives enters the library, delivered only if it improves held-out accuracy. When code carries the reasoning, debugging code becomes debugging reasoning.

In summary, our contributions are threefold:
\begin{itemize}
  \item \textbf{\bench{}}: a $30$-task benchmark for the program-native \cwi{} regime, in which open code, rather than a provided tool vocabulary, carries the reasoning procedure. It is demanding: in language alone even GPT-5.6-luna stays below $30\%$, and \cwi{} raises it to $43.0\%$.
  \item \textbf{\method{}}: a training-free, two-mode skill-evolution loop that combines observational updates with gold-anchored \methodshort{} over program state and rendered intermediates. Its self-evolved skills lift bare \cwi{} from $32.6\%$ to $55.9\%$ on Qwen3.5-27B and from $43.0\%$ to $66.6\%$ on GPT-5.6-luna.
  \item \textbf{Transferable skills}: the delivered skills transfer as plain text. Injected into a 9B solver, they outperform the skills the 9B evolves for itself ($34.5\%$ vs.\ $30.6\%$); they also transfer across model families, lifting gemma-4-26B from $27.7\%$ to $45.0\%$.
\end{itemize}

\section{Related Work}
\label{sec:related}
\paragraph{Thinking with images.} This line lets an LVLM call tools in the middle of reasoning: V* \citep{vstar} runs guided search, cropping and zooming until a small target becomes legible; Visual Sketchpad \citep{hu2024sketchpad} sketches marks as visual thoughts; DeepEyes \citep{deepeyes} and Pixel Reasoner \citep{pixelreasoner} train the zoom-in with RL. Whatever form the call takes, the tool's job is to hand evidence back to a language reasoner, which produces the answer. The division survives when tools are reached through model-written code: in PyVision \citep{pyvision,pyvisionrl}, Thyme \citep{thyme}, CodeV \citep{codev}, and CodeDance \citep{codedance}, the generated programs largely invoke the same perceptual operations (cropping, rotation, contrast enhancement), and reasoning resumes in language over the returned view; earlier, VisProg \citep{gupta2023visprog} and ViperGPT \citep{suris2023vipergpt} compose pretrained vision APIs, again calling capabilities rather than implementing them. \cwi{} has nothing to call: no tool vocabulary is supplied, the model authors the visual algorithm itself, and the program's output is the answer (\S\ref{sec:bench:regime}). We treat program-native reasoning as the object of non-parametric learning, with executable failures becoming the supervision for persistent skill evolution across tasks.

\paragraph{Benchmarks for agentic visual reasoning.} General-purpose suites such as MMMU \citep{mmmu}, MMStar \citep{mmstar}, MathVista \citep{mathvista}, and MathVerse \citep{mathverse} assess broad multimodal understanding over human-curated question sets. Closer to us, V*Bench \citep{vstar} and HR-Bench \citep{hrbench} probe search and high-resolution perception with no code at all. The tool-use suites test how well an agent drives a toolchain: VisualToolBench \citep{visualtoolbench} tests directed cropping and enhancement, Agentic-MME \citep{agenticmme} pairs image operations with web search, AgentVista \citep{agentvista} covers real-world workflows, VTC-Bench \citep{vtcbench} scores the chaining of a fixed vocabulary of $32$ OpenCV operations against ground-truth trajectories, and TIR-Bench \citep{tirbench} asks for tools built at solve time. What they stress is chiefly perception, retrieval, or arithmetic rather than the algorithm itself: as the audits cited in the introduction show, most of what these suites' agents solve with tools they also solve without them \citep{tooluseaudit,codevision}. \bench{} inverts the bottleneck: every target is defined by a computation whose precision defeats inspection, so tool-free failure reflects missing \emph{computation}, and supplying a bare interpreter, nothing else, recovers it. No existing suite makes the authored program the required, measured object; Section~\ref{sec:bench} lays out the comparison in detail.

\paragraph{Skill and experience evolution.} Reflexion \citep{shinn2023reflexion}, ExpeL \citep{zhao2024expel}, and Voyager \citep{wang2023voyager} turn interaction histories into verbal feedback, cross-task experience, or executable skills; GEPA \citep{gepa}, EvoSkill \citep{evoskill}, and SkillOpt \citep{skillopt2026,skilloptlite2026} optimize prompts or skill artifacts from rollout feedback. These methods primarily target text-dominant QA, search, coding, document, or embodied-agent tasks. XSkill extends training-free skill accumulation to visually grounded multimodal trajectories \citep{xskill}, while SPyCE co-evolves hierarchical visual skills with an RL-trained policy \citep{spyce}. Self-Debug \citep{chen2024selfdebug} uses execution feedback to repair a single program within an episode, in the text-and-code domain and without persistence; \methodshort{} lifts executable debugging to the skill level in the \cwi{} regime: verified repairs persist as portable skills while the solver stays frozen (\S\ref{sec:method}).

\section{Code-with-Image Task Regime and Benchmark}
\label{sec:bench}

\subsection{From thinking with images to Code-with-Image}
\label{sec:bench:regime}
The two paradigms draw the line between reasoning and environment in different places. In thinking with images, the environment supplies operations with fixed semantics (crop, zoom, rotate, search, calculator): each call returns evidence, the model reasons over that evidence in language, and the model produces the answer. In \cwi{}, the environment supplies execution only: the model writes the program, the interpreter holds every intermediate as data (arrays, masks, spectra, fitted parameters), and the final program output is the answer, which the model merely transcribes. What separates them is therefore not the interface but where the answer is produced.

Formally, an episode alternates model outputs and environment returns. Write $M$ for the model, $I$ for the task image(s), $h_{k-1}$ for the interaction history before step $k$, and $\hat{y}$ for the returned answer, with $K$ the final step. In thinking with images, $M$ selects an action $a_k$ from a supplied tool vocabulary $\mathcal{T}$, and the selected tool $T_{a_k}$ turns the image into an observation $o_k$. In \cwi{}, $M$ writes a program $p_k$, and $E$, the sandboxed Python interpreter, runs it on the image and the persistent state $s_{k-1}$ (the variables and arrays left by earlier steps), returning the new state $s_k$ with printed and rendered output $o_k$:
\begin{equation}
\begin{aligned}
\text{thinking with images:}\quad
& a_k\sim M(\cdot\mid h_{k-1}), \qquad o_k=T_{a_k}(I), \qquad\ \, \hat{y}\sim M(\cdot\mid h_K);\\
\text{\cwi{}:}\quad
& p_k\sim M(\cdot\mid h_{k-1}), \qquad (s_k,o_k)=E(p_k;\,s_{k-1},I), \qquad \hat{y}=o_K.
\end{aligned}
\label{eq:tool-vs-cwi}
\end{equation}
The loops differ in the last term. Above, every return $o_k$ is folded into the history and the answer is generated by the model; a calculator is the special case where $T_{a_k}$ evaluates an expression whose operands language already supplied. Below, the persistent state $s_k$ carries the intermediates as data, and the answer is the final program output, the model's tokens authoring $p_k$ and transcribing $o_K$ but never carrying the values in between. This is the program-carried reasoning of PoT and PAL \citep{chen2022pot,gao2023pal}, applied where language cannot even state the operands. A task family lies in the \cwi{} \emph{regime} when, at the scored precision, only \cwi{} episodes succeed: recovering the target takes a multi-step computation over pixel values, and no chain of model-generated tokens can carry it. \bench{} is built to isolate this regime. Every instance hands the solver complete evidence at adequate resolution, with nothing to zoom for and nothing to retrieve; when a tool-free model fails, the only possible cause is that the required computation never happened.

\subsection{Comparison with visual reasoning benchmarks}
\label{sec:bench:compare}

\begin{table}[t!]
\centering
\footnotesize
\setlength{\tabcolsep}{3.5pt}
\renewcommand{\arraystretch}{1.12}
\caption{\textbf{Comparison of \bench{} with representative visual reasoning benchmarks.} \emph{Multi-turn} = the solver acts and observes over multiple rounds; \emph{free code} = models author arbitrary programs rather than call a fixed operation set; \emph{answer by code} = most of the final answers are the output of the model's program, not a claim the model generates; \emph{beyond human} = targets cannot be determined by direct human inspection at the scoring precision and must be computed; \emph{generator}/\emph{splits} = new instances sample without bound, with disjoint train/val/test sets. Counts follow the cited papers; ours lists the released evaluation set (resamplable without bound).}
\label{tab:benchmark-positioning}
\begin{tabular*}{\linewidth}{@{\extracolsep{\fill}}l|ccccccccc@{}}
\toprule
Benchmark & Tasks & QA & Tools & \makecell{Multi-\\turn} & \makecell{Free\\code} & \makecell{Answer\\by code} & \makecell{Beyond\\human} & Generator & Splits \\
\midrule
MMStar \citeyearpar{mmstar}                    & 6 & 1.5K & \no  & \no  & \no  & \no & \no  & \no & \no \\
MathVista \citeyearpar{mathvista}             & 5 & 6.1K & \no  & \no  & \no  & \no & \no  & \no & \no \\
V*Bench \citeyearpar{vstar}                   & 2 & 191 & \yes & \no  & \no  & \no & \no  & \no & \no \\
HR-Bench 4K \citeyearpar{hrbench}             & 6 & 200 & \yes & \no  & \no  & \no & \no  & \no & \no \\
VisualToolBench \citeyearpar{visualtoolbench} & 5 & 1.2K & \yes & \yes & \yes & \no & \no  & \no & \no \\
Agentic-MME \citeyearpar{agenticmme}          & 35 & 418 & \yes & \yes & \yes & \no & \no  & \no & \no \\
VTC-Bench \citeyearpar{vtcbench}              & 9 & 680 & \yes & \yes & \yes & \no & \no  & \no & \no \\
TIR-Bench \citeyearpar{tirbench}              & 13 & 1.2K & \yes & \yes & \yes & \no & \no  & \no & \no \\
\midrule
\textbf{\bench{} (ours)} & \textbf{30} & $\boldsymbol{\infty}$\,{\scriptsize(6.3K)} & \yes & \yes & \yes & \yes & \yes & \yes & \yes \\
\bottomrule
\end{tabular*}
\end{table}

Recent benchmarks already share the interaction machinery: most are tool-equipped and multi-turn, and the strongest also accept free-form code as Table~\ref{tab:benchmark-positioning} shows. But there the code exists to \emph{call tools}---crop, enhance, retrieve---for an answer the model writes itself: what these suites test is tool use. \bench{} differs in what is measured and in how targets come to exist. First, the answer is produced by code. In prior suites the graded object is an answer the model generates, however assisted; code, where allowed, mostly serves to expose evidence for a language reasoner. In \bench{} the target is recoverable only as the output of an authored computation, so grading the answer grades the program behind it.

Second, the targets are beyond human ability: each is a precise quantity that no amount of inspection determines, so human annotation is not only unnecessary but impossible. This lifts the constraint that shapes most existing suites, where hand-labeling is expensive and released sets therefore stay small and fixed. Third, correctness and scale come from the generators. Every instance is produced by a program that constructs the image from a known target, so the label is correct by construction rather than by verification; with no human annotator in the loop, labeled instances can be generated without bound. Finally, because most benchmarks offer no data on which a method may learn or tune, \bench{} ships disjoint train, validation, and test splits, with source photographs guaranteed non-overlapping across splits (a zero-intersection audit); this is what makes procedure learning measurable, not only procedure execution.

\subsection{Benchmark construction}
\label{sec:bench:construction}
\begin{figure}[t]
\centering
\includegraphics[width=\linewidth]{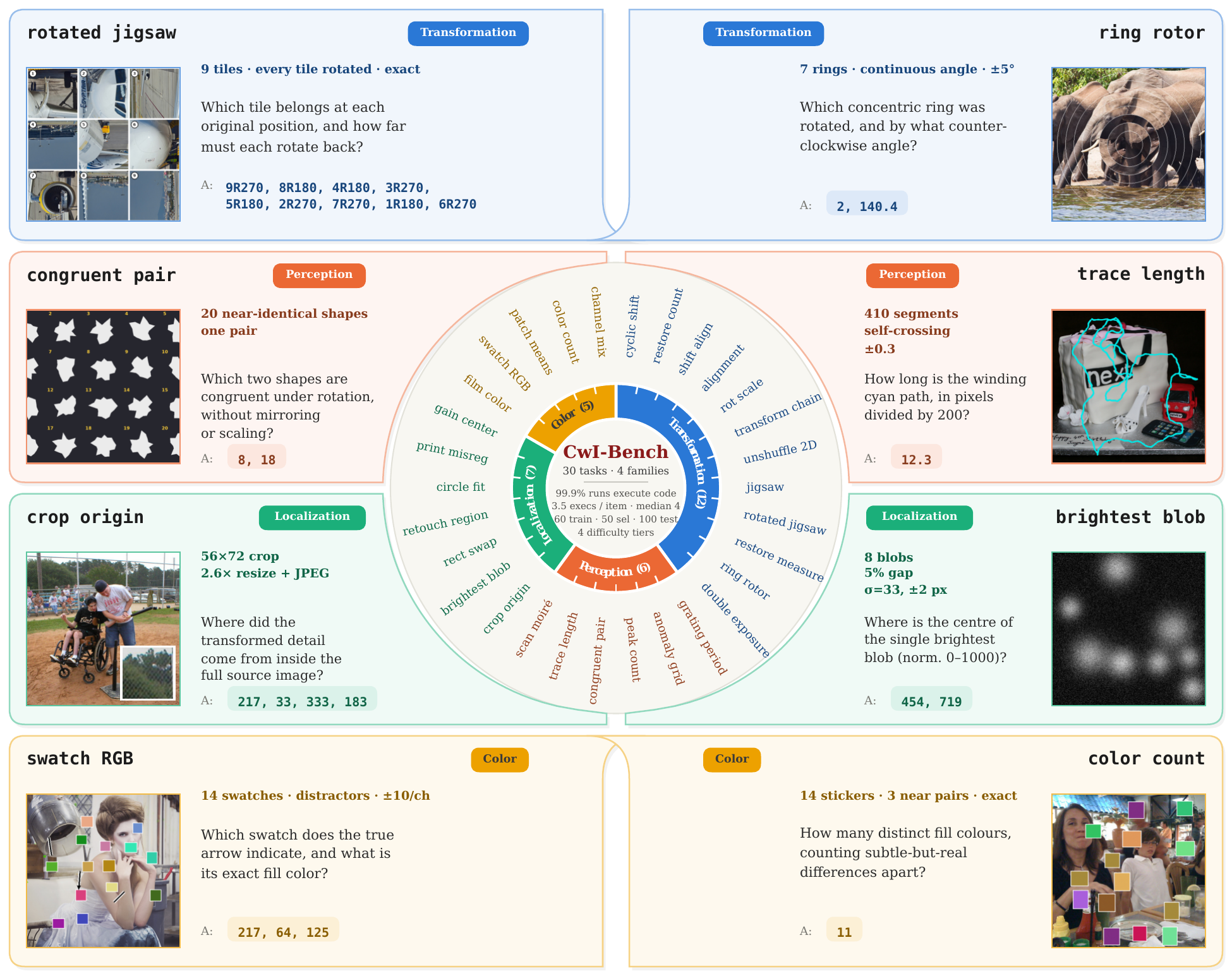}
\caption{\textbf{Benchmark overview.} Each row pairs two representative held-out examples from one semantic family (difficulty controls, question, and constructive ground-truth answer); the waist cards curve around the central wheel. Wheel: all $30$ task distributions, named and colored by family (inner ring $=$ the four families, angle $\propto$ task count); the hub reports the key statistics --- $99.9\%$ of bare-\cwi{} test trajectories execute code ($3.5$ executions per item, median $4$), $60/50/100$ train/val/test items per task, four difficulty tiers. Task cards for all $30$ distributions are cataloged with worked examples in Appendix~\ref{app:tasks}.}
\label{fig:benchmark-overview}
\end{figure}

We operationalize these requirements as a program-first benchmark: each of the $30$ task families is induced by a latent visual computation, and each instance asks the solver to recover that computation's target from the saved pixels. No task-specific API or reference toolchain is provided. The families span four semantic groups, each probing a different kind of authored computation (Figure~\ref{fig:benchmark-overview}). \emph{Transformation and rearrangement} tasks apply a hidden geometric transform to a photograph and ask for its exact parameters, testing whether the solver can invert a transform by searching over it rather than describing it. \emph{Perception and counting} tasks hide weak or near-threshold structure whose extraction requires signal processing, not sharper looking. \emph{Localization} tasks ask for coordinates at a precision that only scanning statistics over the whole array can reach. \emph{Color} tasks ask for exact values and mixing relations that the eye can rank but cannot measure. In several families the natural first implementation harbors a silent fault (an unsigned wrap, an off-by-one convention), so progress requires debugging one's own computation, not choosing a different strategy. Per-task cards with definitions, tolerances, and worked examples for every family are cataloged task by task in Appendix~\ref{app:tasks}.

By construction, these targets resist direct inspection. First, every target is a number, not a thing to name: the eye can tell that a photograph has been rolled or blended, but not by how much, and the scored tolerance (typically two pixels on a $512$-pixel canvas, or a few percent of a continuous value) is finer than visual estimation can reach. Second, any instance that could be guessed by eye is thrown away: each generator checks that the true answer beats its closest visually plausible look-alike by a clear margin, so if an eyeball guess would land within tolerance, the sample never enters the benchmark. Third, many answers are all-or-nothing composites (a full permutation, a tuple of coordinates), so getting most of the answer right still scores zero. Looking harder therefore does not converge on the answer; only executing the right visual algorithm does.

Each task has four difficulty tiers, and ground truth is constructive: an instance is built from its answer rather than annotated after the fact. The generator samples the target quantity first and renders the image from it, so the label is correct by construction; only samples that pass the uniqueness and separability screens above are kept. Training images come from COCO \citep{coco} \texttt{train2017}, and testing exclusively from \texttt{val2017}; an automated audit enforces zero source-image intersection. For each task, reflection sees a $60$-item training slice, skills are selected on a disjoint $50$-item validation set, and every reported number is on $100$ fully held-out test items.

\section{\method}
\label{sec:method}

\begin{figure}[t]
\centering
\includegraphics[width=\linewidth]{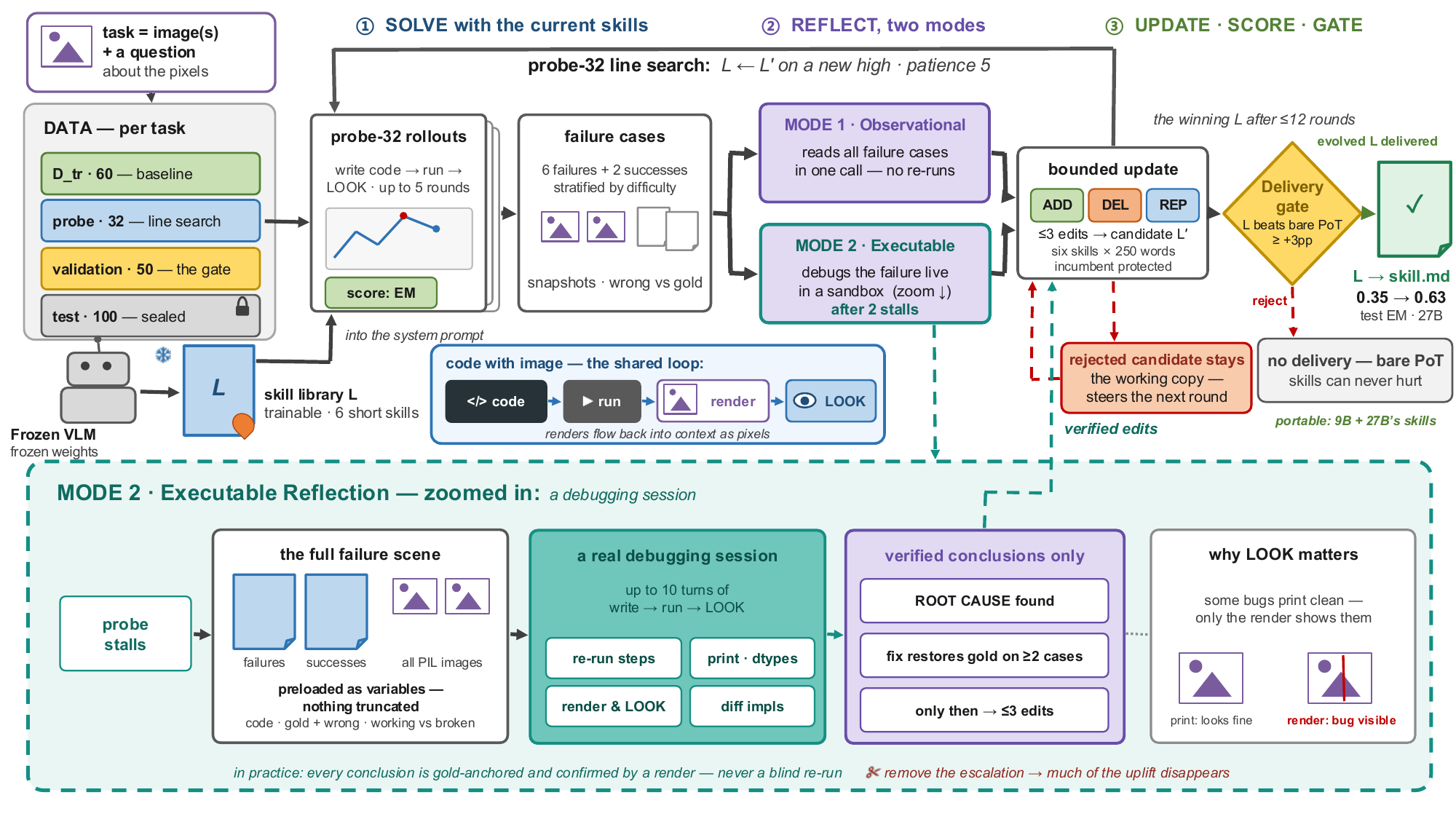}
\caption{\textbf{Overview.} A frozen LVLM solves each task by \cwi{}: the image enters a persistent sandbox as data, and an authored program computes the answer. Failed trajectories drive two reflection modes: observational reflection reads complete multimodal trajectories and proposes bounded edits to a small skill library; once progress stalls, executable reflection re-executes the failures in the sandbox, renders intermediates, and verifies candidate repairs against gold. The edited library is delivered only if it beats the bare solver on a held-out validation split, and at solve time it is injected into the frozen solver's prompt.}
\label{fig:overview}
\end{figure}

\subsection{Setup: learning procedures without weight updates}
\label{sec:method:problem}
\label{sec:method:skills}
Bare \cwi{} gives the model a sandbox in which to execute visual algorithms; it does not tell it what to write. The model knows how to code, but not which algorithm fits the task, which convention the data follows, or which check exposes a silent bug. Reflection pays here because, within a family, failures tend to recur in the same or similar form (a wrong algorithm, an off-by-one, a fragile estimator), so one repair carries across the distribution. A perception task offers no such purchase: a lesson from one image expires with it. In \cwi{} the error lives in the procedure, and so does the repair.

We therefore let the model learn what to write from its own attempts and keep what it learns. Weight updates are the wrong tool: coding ability is not what is missing (and closed models expose no weights), while the missing task knowledge fits in a short text. So the model maintains, for each task, a small library of written procedures, \emph{skills}: it attempts training instances, reflects on failures, edits the library, and keeps an edit only if it improves held-out performance. A skill is plain text: free to apply, inspectable, and portable to a different model. The rest of this section formalizes the loop.

Formally, a task $T$ induces a distribution $\mathcal{D}_T$ over instances $(x,y)$: $x$ bundles the task image(s) with the prompt, and $y$ is the constructively verified answer.
A \emph{skill} is a short model-written text with three fields: \textit{When}, an applicability condition; \textit{Procedure}, numbered operating steps; and \textit{Verify}, which intermediate quantity to render as an image and what correct versus silently broken looks like.
A \emph{library} $L$ is a set of at most $6$ skills of at most $250$ words each; the cap is the regularizer, since a library too small to memorize the training split must compress experience into reusable procedure. The skill library is the object of optimization: the loop edits, scores, and delivers $L$. Each task evolves its own library; the task index stays implicit here.
Solving is the fixed \cwi{} loop, written as the operator \mbox{$\mathcal{A}^{M}_{E}$}: $M$ is the frozen model that authors the programs, $E$ the stateful sandbox that executes them, and $\mathcal{A}$ the bounded act--execute--observe episode they run together. Given the context $c=x\oplus L$ ($x$ with the library appended to its prompt), the solver produces the trajectory
\begin{equation}
\tau \;=\; \big((p_1,s_1,o_1),\ldots,(p_K,s_K,o_K),\hat{y}\big) \;=\; \mathcal{A}^{M}_{E}(c),
\label{eq:operator}
\end{equation}
with $p_k$, $s_k$, $o_k$ the authored program, sandbox state, and execution output of step $k$, as in \eqref{eq:tool-vs-cwi}. The final element $\hat{y}$ is the answer of this episode; when the library matters we write it $\hat{y}_L$. The presence of $L$ is the only difference between bare \cwi{} and \cwi{} with a self-evolved or transferred skill.
Each item is scored $0/1$: exactly for discrete answers (permutations, counts, index sets, transform tuples), and within the task-defined tolerance for continuous ones (pixels, angles, RGB values, percentages); a task's \emph{accuracy} is the mean over its items. With $s_T(\hat y,y)\in\{0,1\}$ this per-item score, learning maximizes $J_T(L)$, the expected accuracy of the solver equipped with library $L$ over the task distribution:
\begin{equation}
J_T(L) \;:=\; \mathbb{E}_{(x,y)\sim\mathcal{D}_T}
\big[\,s_T\big(\hat{y}_L,y\big)\,\big],
\qquad
L^{\star} \;=\; \argmax_{L} \; J_T(L),
\label{eq:objective}
\end{equation}
under two constraints: the model stays frozen, and the only supervision is the solver's own rollouts on a training split $D_{\mathrm{tr}}$, drawn from $\mathcal{D}_T$ disjointly from the test items, together with their gold labels.

\subsection{Skill evolution through self-reflection}
\label{sec:method:rule}
The loop improves the library with two reflection modes, a cheap one that reads failures and an expensive one that re-runs them in a sandbox; it starts with the first and switches to the second only after several rounds bring no improvement (Algorithm~\ref{alg:loop}).

\begin{algorithm}[t]
\caption{\textsc{SkillLearning}: one task, frozen model (full version: Appendix~\ref{app:algo})}
\label{alg:loop}
\begin{algorithmic}[1]
\REQUIRE frozen model $M$; sandbox $E$; training split $D_{\mathrm{tr}}$ with gold labels; probe split $D_{\mathrm{probe}}$ and validation split $D_{\mathrm{val}}$ (scores $\hat{J}_{\mathrm{probe}}$, $\hat{J}_{\mathrm{val}}$); patience $p$; delivery margin $\gamma$
\ENSURE A delivered skill library $L^{\dagger}$ for solve-time injection, or nothing (task keeps the bare solver)
\STATE solve every $x \in D_{\mathrm{tr}}$ with bare \cwi{}; cache the trajectories as $\mathcal{T}_0$; library $L_0 \leftarrow \varnothing$; incumbent (best-so-far) $L^{\dagger} \leftarrow L_0$
\FOR{round $r = 1, 2, \dots$ (fixed budget; early stop on stagnation)}
   \STATE $B_r \leftarrow$ stratified minibatch from $\mathcal{T}_0$: failures from every still-failing tier $+$ same-task successes
   \STATE $\alpha_r \leftarrow$ attendance record: each recent round's probe score with the skills then present
   \IF{$\hat{J}_{\mathrm{probe}}$ stalled for $p$ consecutive rounds}
      \STATE edits $\leftarrow R_{\textsc{exec}}(L_{r-1};B_r,\alpha_r)$ \COMMENT{$m_r{=}\textsc{exec}$: sandbox experiments; repairs tested against gold}
   \ELSE
      \STATE edits $\leftarrow R_{\textsc{obs}}(L_{r-1};B_r,\alpha_r)$ \COMMENT{$m_r{=}\textsc{obs}$: one read-only call over the batch}
   \ENDIF
   \STATE $L_r \leftarrow \Pi_{\mathbb{L}}(L_{r-1}, \text{edits})$ \COMMENT{\textsc{add}/\textsc{replace}/\textsc{delete} within capacity; entries of $L^{\dagger}$ never deleted}
   \STATE \textbf{if} $\hat{J}_{\mathrm{probe}}(L_r)$ best so far \textbf{then} $L^{\dagger} \leftarrow L_r$
\ENDFOR
\STATE evaluate $\hat{J}_{\mathrm{val}}(L^{\dagger})$ on the held-out validation split (runner-up if the incumbent misses)
\STATE \textbf{deliver} iff $\hat{J}_{\mathrm{val}}(L^{\dagger}) \ge \hat{J}_{\mathrm{val}}(\varnothing) + \gamma$ (\eqref{eq:gate}); 
\end{algorithmic}
\end{algorithm}

\paragraph{Self-improvement system.}
The loop that optimizes \eqref{eq:objective} is specified by the tuple
\begin{equation*}
\mathcal{S}=\big(M,\;E,\;\mathbb{L},\;R_{\textsc{obs}},\;R_{\textsc{exec}},\;\{\hat{J}_{\mathrm{probe}},\hat{J}_{\mathrm{val}}\},\;\gamma\big),
\end{equation*}
where $M$ is a frozen LVLM, $E$ a stateful code sandbox that returns program text output and every rendered image, $\mathbb{L}$ the space of admissible skill libraries (capacity-capped sets of When/Procedure/Verify triples), $R_{\textsc{obs}}$ the observational reflection call (read a failure batch, propose edits), $R_{\textsc{exec}}$ the \methodshort{} session (\S\ref{sec:method:evr}), $\hat{J}_{\mathrm{probe}}$ and $\hat{J}_{\mathrm{val}}$ empirical estimators of $J_T$ on a train-side probe and a held-out validation split (\S\ref{sec:method:delivery}), and $\gamma$ a delivery margin requiring a delivered library to beat bare \cwi{} by at least $\gamma$ on validation.
No component of $\mathcal{S}$ updates any weight.

$J_T$ admits no gradient: libraries are discrete text and the solver is frozen, so the only oracle is evaluation.
We therefore implement self-reflection as zeroth-order (ZO) skill optimization, in the lineage of classical methods that estimate update directions from function evaluations alone \citep{spall1992spsa,ghadimi2013zo,nesterov2017random}; forward-only adaptation is viable even for language-model weights \citep{mezo}, and recent analyses cast reflection-style updates as ZO operators on text \citep{skilloptlite2026, zoskill2026, skillopt2026}.
Round $0$ solves the training split with the empty library $L_0=\varnothing$ and caches the trajectories as $\mathcal{T}_0$.
Round $r$ then updates the library, $L_{r-1} \to L_r$, by the textual analogue of a single zeroth-order ascent step:
\begin{equation}
\underbrace{\theta_{r}=\theta_{r-1}+\eta\,\hat{g}(\theta_{r-1};B_r)}_{\text{ZO ascent}}
\;\Longleftrightarrow\;
\underbrace{L_{r}=\Pi_{\mathbb{L}}\!\big(L_{r-1},\,R_{m_r}(L_{r-1};B_r,\alpha_r)\big)}_{\text{reflective update}},
\label{eq:update}
\end{equation}
On the left, classical ZO ascent moves $\theta_{r-1}$ to $\theta_r$ along a direction estimated from function evaluations alone. On the right, the counterpart of $\hat{g}$ is the edit set produced by the reflection call $R_{m_r}(L_{r-1};B_r,\alpha_r)$, which reads the current library $L_{r-1}$ together with two kinds of evidence. The first is a stratified minibatch $B_r\subset\mathcal{T}_0$: failures from every still-failing difficulty tier, joined by same-task successes. The contrast borrows the idea behind two-sided differencing: what the two kinds of trajectory share cancels out, and what separates success from failure stands out. The second input is the attendance record $\alpha_r$: a log of the recent rounds, each line pairing that round's probe score with the set of skills then in the library, so that a persistent rise or fall can be traced to the entries present for it rather than to one round's noise. From these, $R_{m_r}$ emits bounded edits, and $\Pi_{\mathbb{L}}$ executes them (\textsc{add}/\textsc{replace}/\textsc{delete}) under the capacity cap, never deleting an entry of the incumbent $L^{\dagger}$, the best library so far by probe score.
Concretely: $m_r=\textsc{obs}$ by default, and $m_r=\textsc{exec}$ once $\hat{J}_{\mathrm{probe}}$ has not improved for $p$ consecutive rounds.
The two sides differ in the oracle. Classical ZO receives one scalar per query and must perturb blindly; an agentic oracle returns a readable trajectory, so the update is extracted from evidence rather than from random perturbation \citep{skilloptlite2026,zoskill2026}. Executable reflection makes the oracle \emph{re-enterable}: $R_{\textsc{exec}}$ reopens the failed evaluation and measures the direction by experiments on that computation.

For a delivered skill to generalize rather than memorize, its contribution must stay stable to any single training instance; we keep this per-instance sensitivity small through the stratified batch, a syntactic contract forbidding instance-specific values in any entry, and the capacity cap on $\mathbb{L}$.

\begin{table}[t]
\centering
\small
\caption{\textbf{Benchmark evaluation.} Mean test accuracy by task family, all $30$ tasks (per-item $0/1$ accuracy with per-task tolerances, App.~\ref{app:tasks}). \emph{Instruct}/\emph{Thinking} are tool-free CoT with the model's thinking mode off vs.\ on; \emph{think w/ image} rows evaluate upstream tool agents \citep{deepeyes,deepeyesv2,thyme} under their own native protocols on the same items. \textbf{Bold}: best regime per model. Protocol notes: Appendix~\ref{app:pertask}.}
\label{tab:overview}
\setlength{\tabcolsep}{2.5pt}
\resizebox{\linewidth}{!}{%
\begin{tabular}{llccccc}
\toprule
Model & Method & Transformation & Perception & Localization & Color & Overall \\
\midrule
DeepEyes-7B & Think w/ image & 1.8 & 3.8 & 0.1 & 2.4 & 1.9 \\
Thyme-RL & Think w/ image & 1.5 & 4.3 & 0.6 & 2.0 & 1.9 \\
DeepEyesV2-RL & Think w/ image & 1.3 & 4.3 & 1.1 & 3.0 & 2.2 \\
\midrule
\multirow{3}{*}{Qwen3.5-9B} & Instruct & 4.9 & 13.8 & 6.0 & 6.4 & 7.2 \\
 & Thinking & \textbf{9.5} & \textbf{19.3} & 7.3 & 7.4 & 10.6 \\
 & \cwi{} & 8.9 & 16.7 & \textbf{8.6} & \textbf{13.2} & \textbf{11.1} \\
\midrule
\multirow{3}{*}{Qwen3.5-35B-A3B} & Instruct & 6.3 & 19.5 & 5.6 & 9.2 & 9.3 \\
 & Thinking & 10.4 & 24.5 & 7.6 & 10.0 & 12.5 \\
 & \cwi{} & \textbf{22.6} & \textbf{35.5} & \textbf{18.9} & \textbf{24.6} & \textbf{24.6} \\
\midrule
\multirow{3}{*}{Qwen3.5-27B} & Instruct & 6.1 & 18.3 & 4.6 & 8.8 & 8.6 \\
 & Thinking & 12.6 & 24.8 & 11.6 & 10.2 & 14.4 \\
 & \cwi{} & \textbf{27.9} & \textbf{47.0} & \textbf{28.3} & \textbf{32.4} & \textbf{32.6} \\
\midrule
\multirow{3}{*}{gemma-4-12B} & Instruct & 4.5 & 17.3 & 12.7 & 7.2 & 9.4 \\
 & Thinking & 6.1 & \textbf{20.2} & 12.4 & 7.6 & 10.6 \\
 & \cwi{} & \textbf{6.6} & 19.0 & \textbf{18.0} & \textbf{26.6} & \textbf{15.1} \\
\midrule
\multirow{3}{*}{gemma-4-26B-A4B} & Instruct & 6.8 & 17.5 & 11.4 & 11.4 & 10.8 \\
 & Thinking & 8.2 & 17.2 & 14.9 & 8.0 & 11.5 \\
 & \cwi{} & \textbf{16.2} & \textbf{40.2} & \textbf{29.6} & \textbf{37.4} & \textbf{27.7} \\
\midrule
\multirow{3}{*}{GPT-5.6-luna} & Instruct & 11.9 & 16.2 & 10.3 & 15.0 & 12.9 \\
 & Thinking & 29.8 & 28.0 & 16.4 & 22.0 & 25.0 \\
 & \cwi{} & \textbf{37.0} & \textbf{38.2} & \textbf{43.7} & \textbf{62.0} & \textbf{43.0} \\
\bottomrule
\end{tabular}%
}

\end{table}

\paragraph{Observational reflection.}
The default mode is one read-only call. Here $R_{\textsc{obs}}(L_{r-1};B_r,\alpha_r)\sim M(\cdot\mid L_{r-1},B_r,\alpha_r)$ reads the batch (prompts, code, outputs, and every rendered image) and emits bounded edit operations. Reading is enough for strategy-level faults: choosing a different objective, adding a Verify clause, or deleting a misleading entry.

\paragraph{Executable reflection.}
\label{sec:method:evr}
Some failure causes cannot be read off a trajectory. The transcript shows what the code printed and what the model answered, but a silent implementation fault leaves its signature in what was never printed (an intermediate array, a dtype, an off-by-one convention), and no amount of re-reading surfaces it. Such causes can only be found by running the computation again and experimenting on it. When observational reflection stalls, the loop therefore switches from reading its failures to intervening on them: a sandbox $E'$ is preloaded with the failure batch $B_r$: each case's images, gold answer, wrong answer, and the code the solver originally ran. The model then investigates step by step. At step $t$ it writes a piece of investigation code $q_t$ (re-run the old code, print values and dtypes, render an intermediate, or test a candidate repair against gold); the sandbox returns the output $o_t$; and the pair joins a growing session transcript $\mathcal{E}_t$ of everything tried so far, which starts as the failure batch itself:
\begin{equation}
\mathcal{E}_0=B_r, \qquad
q_t \sim M(\cdot \mid L_{r-1},\, \mathcal{E}_{t-1}), \qquad
o_t = E'(q_t), \qquad
\mathcal{E}_t = \mathcal{E}_{t-1}\cup\{(q_t,o_t)\}.
\label{eq:evr}
\end{equation}
The session ends when the model judges a cause verified and emits skill edits, or at a step budget $T_{\max}$, where one final call forces the edits out. This is $R_{\textsc{exec}}=\mathcal{A}^{M}_{E'}$: the same \cwi{} operator, pointed at the model's own past failures. Because the program carries the reasoning state, these experiments intervene on the failed computation itself; where code merely assists a language reasoner, reflection can only critique a textual description of what went wrong.
Edits are admitted only after confirmation on multiple independent failing cases.

\subsection{Validation and delivery}
\label{sec:method:delivery}
The loop sees two scores, and the test split is never one of them. A fixed, difficulty-balanced \emph{probe} $D_{\mathrm{probe}}\subset D_{\mathrm{tr}}$ steers the search: every round is scored by its accuracy $\hat{J}_{\mathrm{probe}}$, the best library so far is protected as the incumbent $L^{\dagger}$, and the search runs non-greedily with early stopping and restarts. A held-out \emph{validation} split $D_{\mathrm{val}}$, with score $\hat{J}_{\mathrm{val}}$, is read only at delivery: the incumbent (then the runner-up) must clear the gate against bare \mbox{\cwi{}}, the empty library $\varnothing$:
\begin{equation}
\text{deliver } L^{\dagger} \;\iff\; \hat{J}_{\mathrm{val}}(L^{\dagger}) \;\ge\; \hat{J}_{\mathrm{val}}(\varnothing) + \gamma.
\label{eq:gate}
\end{equation}
If both miss, nothing is delivered and the task keeps bare \cwi{}; among restarts, the one with the best validation score is kept. What is delivered is the library itself; the per-task tables call it the task's delivered skill. The gate is a validation-side safeguard, not a guarantee against test regression.

\section{Experiments}
\label{sec:exp}
\subsection{Protocol}
The experiments proceed in two parts. Part one evaluates frozen models on \bench{}, with no learning involved: Qwen3.5-9B, Qwen3.5-27B, Qwen3.5-35B-A3B, gemma-4-12B, gemma-4-26B-A4B, and GPT-5.6-luna, plus the three think-with-image agents under their own native protocols. Every arm answers the same held-out test-100 items per task; each frozen model is scored in three regimes under identical item slices and scorers: tool-free CoT, shown in tables as \emph{Instruct} (thinking mode off) and \emph{Thinking} (the model's native thinking mode), and bare \cwi{}, the loop of \S\ref{sec:bench:regime}. Scoring is the per-item $0/1$ accuracy defined above (exact for discrete answers, within the task-defined tolerance for continuous ones; per-task tolerances in App.~\ref{app:tasks}), and all numbers are recomputed from archived raw trajectories.
Part two runs the self-improvement loop of \S\ref{sec:method} on four executors, one shared recipe on every task with no per-task tuning: Qwen3.5-9B, Qwen3.5-27B, gemma-4-26B-A4B, and GPT-5.6-luna.
Loop constants shared by every executor: $|D_{\mathrm{tr}}|{=}60$/task, $|D_{\mathrm{probe}}|{=}32$ ($8$/tier), $|D_{\mathrm{val}}|{=}50$ (stratified), patience $p{=}5$, $T_{\max}{=}10$ \methodshort{} steps, $\gamma{=}3$pp, libraries $\le6$ entries $\times\,250$ words, and rollouts $\le5$ solve steps under a $40{,}960$-token budget. Round and restart budgets differ by family and are cataloged with transport and decoding in Appendix~\ref{app:config}.
At solve time the delivered skills enter the same loop; undelivered tasks run the bare solver.

\subsection{Benchmark evaluation}

Table~\ref{tab:overview} summarizes the grid; per-task tables are in Appendix~\ref{app:pertask}.
The regimes separate cleanly at every scale, in both open families and the closed model.
In Instruct mode, tool-free CoT stays between $7\%$ and $13\%$ for all six frozen models, and thinking mode lifts it to at most $25\%$: these targets sit beyond what language-side reasoning recovers, however strong.
Under \cwi{}, where the environment supplies a general-purpose interpreter and nothing else (no visual API, no reference toolchain), the same frozen models rise at every scale ($8.6\%\!\to\!32.6\%$ on 27B, $12.9\%\!\to\!43.0\%$ on luna); $99.9\%$ of 27B \cwi{} trajectories execute code ($3.5$ executions per instance), so code carries the solution rather than assisting a language reasoner. The medium outweighs scale: with a bare interpreter, a mid-size open model sits at the frontier model's thinking-mode level.

\subsection{Self-evolved skills}
\begin{figure}[t]
\centering
\includegraphics[width=\linewidth]{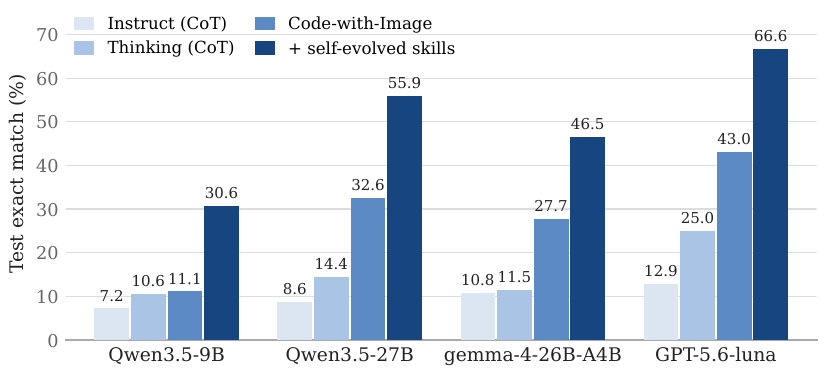}
\caption{\textbf{Self-evolved skills.} Overall test accuracy for the four executors that ran the evolution loop: tool-free CoT (Instruct, Thinking), bare \cwi{}, and \cwi{} with delivered skills (tasks without one fall back to bare \cwi{}). Per-task detail: Appendix~\ref{app:pertask}.}
\label{fig:skillbars}
\end{figure}

Self-evolved skills raise bare \cwi{} for all four executors: $11.1\%\!\to\!30.6\%$ on 9B, $27.7\%\!\to\!46.5\%$ on gemma-26B, $32.6\%\!\to\!55.9\%$ on 27B, and $43.0\%\!\to\!66.6\%$ on luna (Figure~\ref{fig:skillbars}).
On two deterministic tasks the delivered procedure is essentially complete (cyclic shift $100\%$, alignment $97\%$ on 27B test), and no delivered luna skill regresses on test.
The gains are largest where the answer hinges on getting an implementation exactly right, and on those tasks each winning skill was authored inside an executable-reflection session (per-task results: Appendix~\ref{app:pertask}); with its self-evolved skills the open 27B comes within $11$pp of the frontier model's own evolved accuracy.
The reflection loop, then, is not a property of a strong reasoner: whatever the executor's own reasoning strength, the same recipe converts its failures into procedures it can run.

\subsection{Transferred \cwi{} procedures}
Skills are plain text and need not stay with their author; Table~\ref{tab:transfer9b} tests transfer across scale, across model families, and down from the frontier.
Across scale, the 27B model's per-task delivered skills raise the 9B solver from $11.1\%$ to $34.5\%$: above the 9B's own self-evolved skills on every family ($30.6\%$ overall) and past the bare 27B itself ($32.6\%$), with the gain concentrated on Transformation (transferred $38.4\%$ vs.\ self-written $30.1\%$), where procedures are most algorithmic.
Across families, the same library injected into gemma-4-26B-A4B over its native tool-calling transport lifts bare \cwi{} from $27.7\%$ to $45.0\%$, within $1.5$pp of gemma's own evolution ($46.5\%$) and ahead of it on Transformation: nothing the skills encode is specific to their author's family.
Transfer down from the frontier helps, but helps least: luna-written skills lift every recipient over bare \cwi{} ($40.3\%$ on 27B, $17.9\%$ on 9B, $40.8\%$ on gemma) yet land below both the recipient's own evolution and the 27B-written library.
The reason is co-adaptation: a procedure is shaped by what its author can execute, and frontier-written procedures assume a coding robustness that smaller executors lack, so steps that run cleanly for luna fail silently in weaker hands. What transfers is the algorithm; how far it transfers is set by the recipient's ability to run it.

\begin{table}[t]
\centering
\small
\caption{\textbf{Transferred \cwi{} procedures across scale and family} (test-100 family means; accuracy). Tasks without the indicated delivered skill fall back to that solver's bare \cwi{}; \textbf{bold} marks each solver's best regime in every column of the table.}
\label{tab:transfer9b}
\setlength{\tabcolsep}{5pt}
\resizebox{\linewidth}{!}{%
\begin{tabular}{llccccc}
\toprule
Solver & Regime & Transf. & Percep. & Local. & Color & Overall \\
\midrule
\multirow{3}{*}{Qwen3.5-27B} & \cwi{} (no skill) & 27.9 & 47.0 & 28.3 & 32.4 & 32.6 \\
 & \cwi{} + self-evolved & \textbf{54.0} & \textbf{63.8} & \textbf{66.0} & \textbf{36.6} & \textbf{55.9} \\
 & \cwi{} + luna-transferred & 42.5 & 55.0 & 35.1 & 24.4 & 40.3 \\
\midrule
\multirow{4}{*}{Qwen3.5-9B} & \cwi{} (no skill) & 8.9 & 16.7 & 8.6 & 13.2 & 11.1 \\
 & \cwi{} + self-evolved & 30.1 & 32.3 & 37.1 & 20.4 & 30.6 \\
 & \cwi{} + 27B-transferred & \textbf{38.4} & \textbf{33.3} & \textbf{38.1} & \textbf{21.2} & \textbf{34.5} \\
 & \cwi{} + luna-transferred & 23.8 & 21.7 & 8.3 & 12.6 & 17.9 \\
\midrule
\multirow{4}{*}{gemma-4-26B-A4B} & \cwi{} (no skill) & 16.2 & 40.2 & 29.6 & 37.4 & 27.7 \\
 & \cwi{} + self-evolved & 33.7 & \textbf{65.7} & \textbf{56.9} & \textbf{39.8} & \textbf{46.5} \\
 & \cwi{} + 27B-transferred & \textbf{36.5} & 57.0 & 54.0 & 38.6 & 45.0 \\
 & \cwi{} + luna-transferred & 36.2 & 53.3 & 39.3 & 38.8 & 40.8 \\
\bottomrule
\end{tabular}%
}

\end{table}

\begin{table}[t]
\centering
\footnotesize
\caption{\textbf{Ablating executable reflection.} The w/o exec arm replaces escalated rounds with further observational reflection under an otherwise identical loop ($3$ seeds per task). Uplift is the accuracy gain over bare \cwi{} on the validation split, in percentage points; no-delivery runs count $0$. \emph{Mean uplift} averages the three seeds; \emph{best-seed uplift} keeps, per task, the seed with the best validation score (the deployment rule, \S\ref{sec:method:delivery}). \emph{Skills delivered} is the share of the $30\times3$ task$\times$seed runs that deliver a skill.}
\label{tab:ablnoexec}
\begin{tabular}{llccccc}
\toprule
Metric & Arm & Transf. & Percep. & Local. & Color & Overall \\
\midrule
\multicolumn{7}{l}{\textbf{Qwen3.5-27B}} \\
\midrule
\multirow{3}{*}{Mean uplift (pp)} & w/o exec & +7.3 & \textbf{+20.7} & +28.9 & +0.0 & +13.8 \\
 & full & \textbf{+20.2} & +17.3 & \textbf{+32.7} & \textbf{+3.5} & \textbf{+19.7} \\
 & $\Delta$ & +12.9 & -3.3 & +3.8 & +3.5 & +6.0 \\
\midrule
\multirow{3}{*}{Best-seed uplift (pp)} & w/o exec & +15.0 & \textbf{+29.0} & +36.6 & +0.0 & +20.3 \\
 & full & \textbf{+29.7} & +20.7 & \textbf{+41.7} & \textbf{+10.4} & \textbf{+27.5} \\
 & $\Delta$ & +14.7 & -8.3 & +5.1 & +10.4 & +7.1 \\
\midrule
\multirow{3}{*}{Skills delivered (\%)} & w/o exec & 55.6 & \textbf{83.3} & 85.7 & 0.0 & 58.9 \\
 & full & \textbf{75.0} & 72.2 & \textbf{95.2} & \textbf{20.0} & \textbf{70.0} \\
 & $\Delta$ & +19.4 & -11.1 & +9.5 & +20.0 & +11.1 \\
\midrule
\multicolumn{7}{l}{\textbf{Qwen3.5-9B}} \\
\midrule
\multirow{3}{*}{Mean uplift (pp)} & w/o exec & +8.1 & +11.4 & +11.8 & +5.6 & +9.2 \\
 & full & \textbf{+12.2} & \textbf{+12.3} & \textbf{+16.4} & \textbf{+6.1} & \textbf{+12.2} \\
 & $\Delta$ & +4.1 & +0.9 & +4.6 & +0.5 & +3.0 \\
\midrule
\multirow{3}{*}{Best-seed uplift (pp)} & w/o exec & +10.2 & +19.0 & +18.3 & \textbf{+8.8} & +13.6 \\
 & full & \textbf{+21.3} & \textbf{+21.3} & \textbf{+25.7} & +7.6 & \textbf{+20.1} \\
 & $\Delta$ & +11.2 & +2.3 & +7.4 & -1.2 & +6.5 \\
\midrule
\multirow{3}{*}{Skills delivered (\%)} & w/o exec & 33.3 & 72.2 & 38.1 & 40.0 & 43.3 \\
 & full & \textbf{36.1} & \textbf{77.8} & \textbf{57.1} & 40.0 & \textbf{50.0} \\
 & $\Delta$ & +2.8 & +5.6 & +19.0 & +0.0 & +6.7 \\
\bottomrule
\end{tabular}

\end{table}

\subsection{Ablating executable reflection}
We re-run the full evolution loop with executable reflection removed entirely (w/o exec): stalled rounds run further observational reflection instead, on all $30$ tasks $\times$ $3$ seeds ($90$ paired runs per model).
Table~\ref{tab:ablnoexec} reports validation-set uplift over bare \cwi{}.
Removing intervention costs $6.0$pp of the $19.7$pp mean uplift ($7.1$pp at the per-task best seed used for deployment); the interventional share of the total reflection gain stays within a quarter to a third across executors ($6.0/19.7\!=\!30\%$ on 27B vs.\ $3.0/12.2\!=\!24\%$ on 9B).
The loss concentrates on Transformation ($12.9$pp; delivery $27/36\!\to\!20/36$), where faults hide in intermediate state; Localization loses $3.8$pp, while on Perception observational reflection alone is slightly \emph{better} ($-3.3$pp): escalation there displaces observational rounds it did not need.
Where w/o exec wins, the margins are bounded, concentrated where escalation displaces cheap fine-tuning; the intervention-dependent tasks lose $28$--$38$pp at best seed with no recovery path. Reading trajectories refines strategy; the faults that never print require re-execution, and they gate the families where the gains are largest.

\section{Conclusion}
In this paper we introduce \cwi{}, a regime in which a model is given only a general-purpose Python interpreter and must write and run a program that computes the answer, so that the program itself performs the visual reasoning instead of a fixed vocabulary of visual tools, together with \bench{}, a benchmark that isolates this regime with constructive targets beyond inspection precision, unbounded generators, and disjoint splits that make improvement measurable. On top of it we present \method{}, a training-free loop that turns a model's own failed programs into reusable, plain-text skills by re-executing and repairing the silent faults that reading alone cannot reach. Across all six frozen models, reasoning in language alone stays around $7$--$13\%$ and even thinking mode reaches at most $25\%$, whereas a bare Python interpreter lifts accuracy far higher (for example, $12.9\%\!\to\!43.0\%$ on GPT-5.6-luna), and the evolved skills raise it further for every executor while transferring across scale and family to lift models that never evolved them. Because these skills are plain text and improve a frozen model without any weight update, they offer a practical route to accumulating and sharing visual-reasoning procedures across models. When code carries the reasoning, improving a model becomes a matter of debugging its programs, and debugging code becomes debugging reasoning.

\bibliography{iclr2026_conference}
\bibliographystyle{plainnat}

\appendix
\clearpage
\begin{center}
{\Large\bfseries Appendix Table of Contents}
\end{center}
\bigskip
\newcommand{\tocmain}[2]{\noindent\textbf{\hyperref[#2]{#1}} \dotfill\ \pageref{#2}\\[5pt]}
\newcommand{\tocsub}[2]{\noindent\hspace*{1.8em}\hyperref[#2]{#1} \dotfill\ \pageref{#2}\\[3pt]}
\tocmain{A.\enspace Evaluation Configuration}{app:config}
\tocsub{A.1\enspace Shared protocol}{app:config:shared}
\tocsub{A.2\enspace Qwen family}{app:config:qwen}
\tocsub{A.3\enspace gemma family}{app:config:gemma}
\tocsub{A.4\enspace GPT-5.6-luna}{app:config:luna}
\tocmain{B.\enspace Per-Task Results for All Models}{app:pertask}
\tocsub{Qwen3.5-27B}{tab:main}
\tocsub{GPT-5.6-luna}{tab:luna}
\tocsub{Qwen3.5-9B}{tab:qwen9b}
\tocsub{gemma-4-26B-A4B-it}{tab:gemma}
\tocmain{C.\enspace Task Cards and Worked Examples}{app:tasks}
\tocsub{C.1\enspace Transformation \& rearrangement}{app:tasks:transform}
\tocsub{C.2\enspace Perception \& counting}{app:tasks:percep}
\tocsub{C.3\enspace Localization}{app:tasks:local}
\tocsub{C.4\enspace Color}{app:tasks:color}
\tocmain{D.\enspace Reflection Prompts (Verbatim)}{app:prompts}
\tocmain{E.\enspace Exhibits: a Delivered Skill and a Reflection Session}{app:exhibits}
\tocsub{E.1\enspace A delivered multimodal skill, verbatim}{app:exhibits:skill}
\tocsub{E.2\enspace Executable-reflection session excerpts}{app:exhibits:session}
\tocmain{F.\enspace Full Pseudocode}{app:algo}

\clearpage
\section{Evaluation configuration}
\label{app:config}
All arms in the paper share the same held-out items, the same accuracy scorer (exact for
discrete answers, within task-defined tolerances for continuous ones), and the same sandbox implementation. This section records the
complete configuration: the protocol shared by every family, then the per-family
access, transport, and decoding settings.

\subsection{Shared protocol}
\label{app:config:shared}
Every \cwi{} arm runs the bounded act--execute--observe loop of \S\ref{sec:bench:regime}:
at most $5$ solve steps per episode, a forced final answer on the last step, and a
$40{,}960$-token episode budget. The sandbox is a persistent Jupyter kernel per episode
with an isolated working directory; every execution returns stdout plus every rendered
image. The evolution loop uses the constants of \S\ref{sec:exp}: $|D_{\mathrm{tr}}|{=}60$ per task, a difficulty-balanced probe of $32$, a held-out validation split of $50$, patience $5$, $T_{\max}{=}10$ executable-reflection steps, delivery margin $\gamma{=}3$pp, and libraries capped at $6$ entries $\times\,250$ words; the round and restart budgets are per-family, below.
A delivered library is appended to the solver's system prompt at solve time; everything else stays as configured above.

\subsection{Qwen family (Qwen3.5-9B, -27B, -35B-A3B)}
\label{app:config:qwen}
Open weights, served locally with vLLM (bfloat16); Qwen3.5-35B-A3B is evaluated as a
baseline only, under the same settings. \emph{CoT arms}: Instruct decodes with
temperature $0.7$, top-$p$ $0.8$, top-$k$ $20$, presence penalty $1.5$, with the
thinking mode disabled; Thinking switches to the native thinking mode and decodes with
temperature $1.0$, top-$p$ $0.95$. \emph{\cwi{}}: the solver keeps the Instruct
decoding; transport is the native XML tool protocol, the model emitting
\texttt{<tool\_call>} blocks that carry code, with stdout and every rendered image
returned as multimodal content. \emph{Skills}: the delivered library enters the system
prompt, together with two anchor images harvested from the model's own successful runs
and placed before the task images; evolution uses $3$ independent restarts with at most $12$ update rounds each.

\subsection{gemma family (gemma-4-12B, -26B-A4B-it)}
\label{app:config:gemma}
Open weights, served locally with vLLM (bfloat16); gemma-4-12B is evaluated as a
baseline only. All arms decode with the official recommended settings, temperature
$1.0$, top-$p$ $0.95$, top-$k$ $64$; parameters the official configuration does not
specify are not sent. Instruct runs with the thinking mode off (no think token in the
template); Thinking inserts the native think markers. \emph{\cwi{}}: transport is the
model's native function-calling chat template, authored code traveling through the
standard tool-call channel. \emph{Skills}: injected as plain text in the system prompt, with no anchor images. Evolution uses $3$ independent restarts with at most $12$ rounds each, and every task solves with the best of its delivered seeds.

\subsection{GPT-5.6-luna}
\label{app:config:luna}
Closed weights, accessed through the provider's chat-completions API, with sampling parameters at the provider defaults. \emph{CoT arms}: Instruct runs with the
reasoning mode off; Thinking enables the provider's reasoning mode at its default
effort (medium). \emph{\cwi{}}: transport is function tools, with reasoning disabled
inside the tool loop so that authored code, not hidden reasoning, carries the solution.
\emph{Skills}: text-only (no anchor images). The evolution loop deviates from the shared recipe only in scale, using a single restart and $8$ rounds: chiefly, the frontier model's reflection is strong enough that a single restart suffices; the reduced budget also bounds the closed-model API cost. The reflection prompts, edit grammar, minibatch, probe, validation split, and delivery gate are unchanged.

\section{Per-task results for all models}
\label{app:pertask}
Tables~\ref{tab:main}--\ref{tab:gemma} give the per-task results behind Table~\ref{tab:overview} and Figure~\ref{fig:skillbars}, one model per page.

\paragraph{Protocol notes for Table~\ref{tab:overview}.} Models are grouped by family, ordered by bare-\cwi{} accuracy within each family; Qwen3.5-35B-A3B is a baseline-only row (no evolution run). The think-with-image systems ($7$--$8$B) are evaluated under their own native tool protocols (their single supported regime) on the same held-out items with the same scorer. DeepEyes-7B's zoom-only toolset cannot express these computations. All three agents deliver answers on $94$--$98\%$ of items, and their scores concentrate on the few tasks where inspection suffices (counting colors, shapes, or stripes), with zero or near-zero accuracy on every transformation and measurement task.

\begin{table}[H]
\centering
\small
\caption{\textbf{Qwen3.5-27B per-task accuracy (\%).} Held-out test-100, all $30$ tasks. Under \cwi{}, ``No skill'' is the bare solver and ``+ Skill'' uses the delivered self-evolved skill; $\Delta$ is their difference. Skills are delivered only past the $+3$pp validation gate; $^{\circ}$ marks tasks without a delivered skill (+ Skill then repeats the bare solver).}
\label{tab:main}
\setlength{\tabcolsep}{6pt}
\renewcommand{\arraystretch}{1.22}
\begin{tabularx}{\linewidth}{@{}Xrrrrr@{}}
\toprule
\multirow{2}{*}{\textbf{Task}} & \multirow{2}{*}{\textbf{Instruct}} & \multirow{2}{*}{\textbf{Thinking}} & \multicolumn{2}{c}{\textbf{\cwi{}}} & \multirow{2}{*}{\textbf{$\Delta$}} \\
\cmidrule(lr){4-5}
 &  & & \textbf{No skill} & \textbf{+ Skill} & \\
\midrule
\addlinespace[2pt]
\rowcolor{black!6} \multicolumn{6}{l}{\textbf{Transformation \& rearrangement}} \\
cyclic shift & 0.0 & 0.0 & 44.0 & \textbf{100.0} & \gain{56.0} \\
restore count & 8.0 & 40.0 & 24.0 & \textbf{82.0} & \gain{58.0} \\
shift align & 0.0 & 0.0 & 55.0 & \textbf{100.0} & \gain{45.0} \\
alignment & 0.0 & 0.0 & 72.0 & \textbf{97.0} & \gain{25.0} \\
rot scale & 8.0 & 9.0 & 25.0 & \textbf{88.0} & \gain{63.0} \\
transform chain & 37.0 & 52.0 & 73.0 & \textbf{98.0} & \gain{25.0} \\
unshuffle 2D$^{\circ}$ & 3.0 & 19.0 & 8.0 & 8.0 & $0.0$ \\
jigsaw & 0.0 & 0.0 & 22.0 & \textbf{33.0} & \gain{11.0} \\
rotated jigsaw & 1.0 & 1.0 & 2.0 & \textbf{4.0} & \gain{2.0} \\
restore measure & 14.0 & 29.0 & 3.0 & \textbf{7.0} & \gain{4.0} \\
ring rotor & 2.0 & 1.0 & 7.0 & \textbf{28.0} & \gain{21.0} \\
double exposure & 0.0 & 0.0 & 0.0 & \textbf{3.0} & \gain{3.0} \\
\addlinespace[2pt]
\rowcolor{black!6} \multicolumn{6}{l}{\textbf{Perception \& counting}} \\
grating period & 22.0 & 18.0 & 72.0 & \textbf{98.0} & \gain{26.0} \\
anomaly grid & 45.0 & 65.0 & 83.0 & \textbf{95.0} & \gain{12.0} \\
peak count & 20.0 & 44.0 & 37.0 & \textbf{47.0} & \gain{10.0} \\
congruent pair & 9.0 & 10.0 & 67.0 & \textbf{82.0} & \gain{15.0} \\
trace length & 11.0 & 8.0 & 14.0 & \textbf{52.0} & \gain{38.0} \\
scan moir\'e$^{\circ}$ & 3.0 & 4.0 & 9.0 & 9.0 & $0.0$ \\
\addlinespace[2pt]
\rowcolor{black!6} \multicolumn{6}{l}{\textbf{Localization}} \\
crop origin & 9.0 & 11.0 & 32.0 & \textbf{96.0} & \gain{64.0} \\
brightest blob & 0.0 & 3.0 & 28.0 & \textbf{89.0} & \gain{61.0} \\
rect swap & 17.0 & 38.0 & 62.0 & \textbf{89.0} & \gain{27.0} \\
retouch region & 5.0 & 25.0 & 13.0 & \textbf{62.0} & \gain{49.0} \\
circle fit & 0.0 & 0.0 & 17.0 & \textbf{37.0} & \gain{20.0} \\
print misreg & 1.0 & 4.0 & 40.0 & \textbf{82.0} & \gain{42.0} \\
gain center & 0.0 & 0.0 & 6.0 & \textbf{7.0} & \gain{1.0} \\
\addlinespace[2pt]
\rowcolor{black!6} \multicolumn{6}{l}{\textbf{Color}} \\
film color & 1.0 & 5.0 & 39.0 & 29.0 & $-10.0$ \\
swatch RGB & 8.0 & 7.0 & 31.0 & \textbf{37.0} & \gain{6.0} \\
patch means & 1.0 & 4.0 & 34.0 & \textbf{59.0} & \gain{25.0} \\
color count$^{\circ}$ & 34.0 & 35.0 & 58.0 & 58.0 & $0.0$ \\
channel mix$^{\circ}$ & 0.0 & 0.0 & 0.0 & 0.0 & $0.0$ \\
\addlinespace[2pt]
\midrule
\textbf{Mean (30 tasks)} & 8.6 & 14.4 & 32.6 & \textbf{55.9} & \gain{23.3} \\
\bottomrule
\end{tabularx}

\end{table}

\clearpage
\begin{table}[H]
\centering
\small
\caption{\textbf{GPT-5.6-luna per-task accuracy (\%).} The model runs the same loop with its own \cwi{} solver, reflections, and skills. \emph{Instruct} and \emph{Thinking} are tool-free CoT with thinking mode off vs.\ on. $\Delta$ is ``+ Skill'' minus ``No skill''; $^{\circ}$ marks tasks without a delivered skill (gate not passed; falls back to the bare solver).}
\label{tab:luna28}
\label{tab:luna}
\setlength{\tabcolsep}{6pt}
\renewcommand{\arraystretch}{1.22}
\begin{tabularx}{\linewidth}{@{}Xrrrrr@{}}
\toprule
\multirow{2}{*}{\textbf{Task}} & \multirow{2}{*}{\textbf{Instruct}} & \multirow{2}{*}{\textbf{Thinking}} & \multicolumn{2}{c}{\textbf{\cwi{}}} & \multirow{2}{*}{\textbf{$\Delta$}} \\
\cmidrule(lr){4-5}
 &  & & \textbf{No skill} & \textbf{+ Skill} & \\
\midrule
\addlinespace[2pt]
\rowcolor{black!6} \multicolumn{6}{l}{\textbf{Transformation \& rearrangement}} \\
cyclic shift & 73.0 & 69.0 & 81.0 & \textbf{100.0} & \gain{19.0} \\
restore count & 12.0 & 66.0 & 48.0 & \textbf{70.0} & \gain{22.0} \\
shift align & 0.0 & 36.0 & 83.0 & \textbf{98.0} & \gain{15.0} \\
alignment & 0.0 & 11.0 & 74.0 & \textbf{98.0} & \gain{24.0} \\
rot scale & 16.0 & 35.0 & 32.0 & \textbf{100.0} & \gain{68.0} \\
transform chain & 11.0 & 60.0 & 69.0 & \textbf{98.0} & \gain{29.0} \\
unshuffle 2D & 3.0 & 8.0 & 6.0 & \textbf{27.0} & \gain{21.0} \\
jigsaw & 0.0 & 1.0 & 0.0 & \textbf{13.0} & \gain{13.0} \\
rotated jigsaw & 0.0 & 1.0 & 0.0 & \textbf{5.0} & \gain{5.0} \\
restore measure & 26.0 & 48.0 & 32.0 & \textbf{59.0} & \gain{27.0} \\
ring rotor & 2.0 & 2.0 & 5.0 & \textbf{31.0} & \gain{26.0} \\
double exposure & 0.0 & 21.0 & 14.0 & \textbf{17.0} & \gain{3.0} \\
\addlinespace[2pt]
\rowcolor{black!6} \multicolumn{6}{l}{\textbf{Perception \& counting}} \\
grating period$^{\circ}$ & 27.0 & 24.0 & 92.0 & 92.0 & $0.0$ \\
anomaly grid & 16.0 & 72.0 & 63.0 & \textbf{69.0} & \gain{6.0} \\
peak count & 33.0 & 39.0 & 41.0 & \textbf{65.0} & \gain{24.0} \\
congruent pair & 11.0 & 33.0 & 17.0 & \textbf{35.0} & \gain{18.0} \\
trace length & 10.0 & 0.0 & 1.0 & \textbf{63.0} & \gain{62.0} \\
scan moir\'e & 0.0 & 0.0 & 15.0 & \textbf{94.0} & \gain{79.0} \\
\addlinespace[2pt]
\rowcolor{black!6} \multicolumn{6}{l}{\textbf{Localization}} \\
crop origin & 3.0 & 19.0 & 39.0 & \textbf{89.0} & \gain{50.0} \\
brightest blob$^{\circ}$ & 1.0 & 13.0 & 81.0 & 81.0 & $0.0$ \\
rect swap & 34.0 & 13.0 & 47.0 & \textbf{71.0} & \gain{24.0} \\
retouch region$^{\circ}$ & 33.0 & 42.0 & 43.0 & 43.0 & $0.0$ \\
circle fit & 1.0 & 28.0 & 47.0 & \textbf{50.0} & \gain{3.0} \\
print misreg & 0.0 & 0.0 & 19.0 & \textbf{77.0} & \gain{58.0} \\
gain center & 0.0 & 0.0 & 30.0 & \textbf{43.0} & \gain{13.0} \\
\addlinespace[2pt]
\rowcolor{black!6} \multicolumn{6}{l}{\textbf{Color}} \\
film color & 3.0 & 6.0 & 86.0 & \textbf{100.0} & \gain{14.0} \\
swatch RGB & 49.0 & 55.0 & 87.0 & \textbf{91.0} & \gain{4.0} \\
patch means & 4.0 & 13.0 & 64.0 & \textbf{84.0} & \gain{20.0} \\
color count & 17.0 & 31.0 & 64.0 & \textbf{82.0} & \gain{18.0} \\
channel mix & 2.0 & 5.0 & 9.0 & \textbf{52.0} & \gain{43.0} \\
\addlinespace[2pt]
\midrule
\textbf{Mean (30 tasks)} & 12.9 & 25.0 & 43.0 & \textbf{66.6} & \gain{23.6} \\
\bottomrule
\end{tabularx}

\end{table}

\clearpage
\begin{table}[H]
\centering
\small
\caption{\textbf{Qwen3.5-9B per-task accuracy (\%).} All $30$ tasks, using its own self-evolved skills and the best of three restarts. $\Delta$ is ``+ Skill'' minus ``No skill''; $^{\circ}$ marks tasks without a delivered skill (falls back to the bare solver).}
\label{tab:qwen9b}
\setlength{\tabcolsep}{6pt}
\renewcommand{\arraystretch}{1.22}
\begin{tabularx}{\linewidth}{@{}Xrrrrr@{}}
\toprule
\multirow{2}{*}{\textbf{Task}} & \multirow{2}{*}{\textbf{Instruct}} & \multirow{2}{*}{\textbf{Thinking}} & \multicolumn{2}{c}{\textbf{\cwi{}}} & \multirow{2}{*}{\textbf{$\Delta$}} \\
\cmidrule(lr){4-5}
 &  & & \textbf{No skill} & \textbf{+ Skill} & \\
\midrule
\addlinespace[2pt]
\rowcolor{black!6} \multicolumn{6}{l}{\textbf{Transformation \& rearrangement}} \\
cyclic shift & 0.0 & 0.0 & 8.0 & \textbf{84.0} & \gain{76.0} \\
restore count & 11.0 & 21.0 & 6.0 & \textbf{20.0} & \gain{14.0} \\
shift align & 0.0 & 0.0 & 7.0 & \textbf{45.0} & \gain{38.0} \\
alignment & 0.0 & 0.0 & 12.0 & \textbf{84.0} & \gain{72.0} \\
rot scale$^{\circ}$ & 6.0 & 15.0 & 17.0 & 17.0 & $0.0$ \\
transform chain & 23.0 & 50.0 & 41.0 & \textbf{93.0} & \gain{52.0} \\
unshuffle 2D & 6.0 & 13.0 & 2.0 & \textbf{2.0} & $0.0$ \\
jigsaw$^{\circ}$ & 0.0 & 0.0 & 6.0 & 6.0 & $0.0$ \\
rotated jigsaw$^{\circ}$ & 0.0 & 2.0 & 0.0 & 0.0 & $0.0$ \\
restore measure & 10.0 & 11.0 & 6.0 & \textbf{8.0} & \gain{2.0} \\
ring rotor$^{\circ}$ & 3.0 & 2.0 & 2.0 & 2.0 & $0.0$ \\
double exposure$^{\circ}$ & 0.0 & 0.0 & 0.0 & 0.0 & $0.0$ \\
\addlinespace[2pt]
\rowcolor{black!6} \multicolumn{6}{l}{\textbf{Perception \& counting}} \\
grating period & 12.0 & 5.0 & 4.0 & \textbf{15.0} & \gain{11.0} \\
anomaly grid & 35.0 & 53.0 & 60.0 & \textbf{78.0} & \gain{18.0} \\
peak count & 22.0 & 37.0 & 14.0 & \textbf{61.0} & \gain{47.0} \\
congruent pair & 8.0 & 12.0 & 16.0 & \textbf{19.0} & \gain{3.0} \\
trace length & 3.0 & 8.0 & 6.0 & \textbf{21.0} & \gain{15.0} \\
scan moir\'e$^{\circ}$ & 3.0 & 1.0 & 0.0 & 0.0 & $0.0$ \\
\addlinespace[2pt]
\rowcolor{black!6} \multicolumn{6}{l}{\textbf{Localization}} \\
crop origin & 8.0 & 10.0 & 10.0 & \textbf{79.0} & \gain{69.0} \\
brightest blob & 0.0 & 0.0 & 21.0 & \textbf{67.0} & \gain{46.0} \\
rect swap & 23.0 & 29.0 & 18.0 & \textbf{66.0} & \gain{48.0} \\
retouch region & 10.0 & 11.0 & 4.0 & \textbf{22.0} & \gain{18.0} \\
circle fit & 1.0 & 1.0 & 4.0 & \textbf{9.0} & \gain{5.0} \\
print misreg & 0.0 & 0.0 & 2.0 & \textbf{16.0} & \gain{14.0} \\
gain center$^{\circ}$ & 0.0 & 0.0 & 1.0 & 1.0 & $0.0$ \\
\addlinespace[2pt]
\rowcolor{black!6} \multicolumn{6}{l}{\textbf{Color}} \\
film color$^{\circ}$ & 0.0 & 1.0 & 4.0 & 4.0 & $0.0$ \\
swatch RGB$^{\circ}$ & 4.0 & 5.0 & 20.0 & 20.0 & $0.0$ \\
patch means & 1.0 & 1.0 & 16.0 & \textbf{33.0} & \gain{17.0} \\
color count & 27.0 & 29.0 & 26.0 & \textbf{45.0} & \gain{19.0} \\
channel mix$^{\circ}$ & 0.0 & 1.0 & 0.0 & 0.0 & $0.0$ \\
\addlinespace[2pt]
\midrule
\textbf{Mean (30 tasks)} & 7.2 & 10.6 & 11.1 & \textbf{30.6} & \gain{19.5} \\
\bottomrule
\end{tabularx}

\end{table}

\begin{table}[H]
\centering
\small
\caption{\textbf{gemma-4-26B-A4B-it per-task accuracy (\%).} All $30$ tasks, using its own self-evolved skills (best delivered seed, native tool-calling transport). $\Delta$ is ``+ Skill'' minus ``No skill''; $^{\circ}$ marks tasks without a delivered skill (falls back to the bare solver).}
\label{tab:gemma}
\setlength{\tabcolsep}{6pt}
\renewcommand{\arraystretch}{1.22}
\begin{tabularx}{\linewidth}{@{}Xrrrrr@{}}
\toprule
\multirow{2}{*}{\textbf{Task}} & \multirow{2}{*}{\textbf{Instruct}} & \multirow{2}{*}{\textbf{Thinking}} & \multicolumn{2}{c}{\textbf{\cwi{}}} & \multirow{2}{*}{\textbf{$\Delta$}} \\
\cmidrule(lr){4-5}
 &  & & \textbf{No skill} & \textbf{+ Skill} & \\
\midrule
\addlinespace[2pt]
\rowcolor{black!6} \multicolumn{6}{l}{\textbf{Transformation \& rearrangement}} \\
cyclic shift & 0.0 & 0.0 & 11.0 & \textbf{100.0} & \gain{89.0} \\
restore count & 21.0 & 28.0 & 18.0 & 14.0 & $-4.0$ \\
shift align & 0.0 & 0.0 & 53.0 & \textbf{76.0} & \gain{23.0} \\
alignment & 0.0 & 0.0 & 29.0 & \textbf{95.0} & \gain{66.0} \\
rot scale & 0.0 & 4.0 & 2.0 & \textbf{27.0} & \gain{25.0} \\
transform chain$^{\circ}$ & 43.0 & 41.0 & 56.0 & 56.0 & $0.0$ \\
unshuffle 2D & 4.0 & 8.0 & 4.0 & 3.0 & $-1.0$ \\
jigsaw$^{\circ}$ & 0.0 & 0.0 & 1.0 & 1.0 & $0.0$ \\
rotated jigsaw$^{\circ}$ & 0.0 & 0.0 & 0.0 & 0.0 & $0.0$ \\
restore measure & 14.0 & 17.0 & 16.0 & \textbf{20.0} & \gain{4.0} \\
ring rotor$^{\circ}$ & 0.0 & 0.0 & 4.0 & 4.0 & $0.0$ \\
double exposure & 0.0 & 0.0 & 1.0 & \textbf{8.0} & \gain{7.0} \\
\addlinespace[2pt]
\rowcolor{black!6} \multicolumn{6}{l}{\textbf{Perception \& counting}} \\
grating period & 5.0 & 12.0 & 69.0 & \textbf{90.0} & \gain{21.0} \\
anomaly grid & 37.0 & 36.0 & 72.0 & \textbf{82.0} & \gain{10.0} \\
peak count & 49.0 & 38.0 & 45.0 & \textbf{73.0} & \gain{28.0} \\
congruent pair & 6.0 & 2.0 & 30.0 & \textbf{68.0} & \gain{38.0} \\
trace length & 3.0 & 7.0 & 20.0 & \textbf{57.0} & \gain{37.0} \\
scan moir\'e & 5.0 & 8.0 & 5.0 & \textbf{24.0} & \gain{19.0} \\
\addlinespace[2pt]
\rowcolor{black!6} \multicolumn{6}{l}{\textbf{Localization}} \\
crop origin & 0.0 & 6.0 & 25.0 & \textbf{96.0} & \gain{71.0} \\
brightest blob & 10.0 & 4.0 & 44.0 & \textbf{51.0} & \gain{7.0} \\
rect swap & 28.0 & 53.0 & 37.0 & \textbf{89.0} & \gain{52.0} \\
retouch region$^{\circ}$ & 38.0 & 34.0 & 31.0 & 31.0 & $0.0$ \\
circle fit & 4.0 & 6.0 & 40.0 & \textbf{58.0} & \gain{18.0} \\
print misreg & 0.0 & 1.0 & 26.0 & \textbf{59.0} & \gain{33.0} \\
gain center & 0.0 & 0.0 & 4.0 & \textbf{14.0} & \gain{10.0} \\
\addlinespace[2pt]
\rowcolor{black!6} \multicolumn{6}{l}{\textbf{Color}} \\
film color & 2.0 & 1.0 & 26.0 & 24.0 & $-2.0$ \\
swatch RGB$^{\circ}$ & 27.0 & 16.0 & 76.0 & 76.0 & $0.0$ \\
patch means & 8.0 & 3.0 & 27.0 & \textbf{36.0} & \gain{9.0} \\
color count & 19.0 & 18.0 & 58.0 & \textbf{63.0} & \gain{5.0} \\
channel mix$^{\circ}$ & 1.0 & 2.0 & 0.0 & 0.0 & $0.0$ \\
\addlinespace[2pt]
\midrule
\textbf{Mean (30 tasks)} & 10.8 & 11.5 & 27.7 & \textbf{46.5} & \gain{18.8} \\
\bottomrule
\end{tabularx}

\end{table}

\clearpage
\section{Task cards and worked examples}
\label{app:tasks}
Each card's \textbf{Scoring} line states when a prediction is counted correct: it scores $1$ if that condition holds and $0$ otherwise (no partial credit; box and point coordinates are normalized to $0$--$1000$).
\newcommand{\taskcard}[4]{\paragraph{#1.}\textbf{Question.} #2\ \textbf{Construction.} #3\ \textbf{Scoring.} #4}

\subsection{Transformation \& rearrangement}
\label{app:tasks:transform}

\taskcard{cyclic shift}
{The image was produced by cyclically shifting the original picture (pixels wrap around the edges). What are the shift amounts $(dx, dy)$? Tolerance $\pm 2$px, both required.}
{A COCO square is rolled by $(dx,dy)$ drawn per difficulty tier; tiers scale shift range and texture richness.}
{$dx,dy$ each within $\pm 2$px, both required.}

\taskcard{restore count}
{First undo a cyclic shift, then count the shapes entirely in the left half of the restored image. Answer: one integer.}
{Textured shapes composited on a photo background (exact masks recorded), then the frame is rolled.}
{The reported integer must equal the true count.}

\taskcard{shift align}
{Image 2 equals image 1 rotated CCW by $\theta$ about the center, then cyclically shifted by $(dx,dy)$. Report $dx, dy, \theta$ (generation grids 8px / $15^\circ$; scored at $\pm 2$px / $\pm 1.875^\circ$).}
{Circular window (white outside) deliberately removes the wrap-seam shortcut; the two parameters contaminate each other's estimators.}
{$dx,dy$ within $\pm 2$px each and $\theta$ within $\pm 1.875^\circ$ (circular).}

\taskcard{alignment}
{Two images show the same photo; the second is translated. Report $(dx,dy)$, $\pm 2$px.}
{Second view is a cyclic translation of the first; tiers scale offset magnitude.}
{$dx,dy$ each within $\pm 2$px.}

\taskcard{rot scale}
{Image 2 equals image 1 zoomed in by $s\ge 1$ about the center, then rotated CCW by $\theta$ through a circular window. Report $\theta, s$ ($\pm 3^\circ$ and $\pm 0.05$, both required).}
{Rotation/scale applied inside an inscribed disc; tiers move from right-angle/integer-scale to arbitrary parameters.}
{$\theta$ within $\pm 3^\circ$ (circular) and $s$ within $\pm 0.05$, both required.}

\taskcard{transform chain}
{Image 2 was produced from image 1 by some combination of cyclic shifts, right-angle rotations, and mirror flips in unknown order. Report the \emph{canonical form}: \texttt{shift(dx,dy); rot(k); fliph(f)}.}
{Random operation chains; the answer is the unique $D_4\ltimes$shift decomposition (group structure guarantees uniqueness; raw operation order is provably non-identifiable, which we discovered and designed around).}
{The reported $(dx,dy,k,f)$ tuple must match in every field.}

\taskcard{unshuffle 2D}
{Row order and column order were independently shuffled (tiles within a row/column stay together). Report both restoring permutations.}
{Rows and columns permuted with badges on the left/top edges; tiers scale grid size ($2\times3$ to $4\times4$).}
{Both the row and column permutations must match exactly.}

\taskcard{jigsaw}
{The photo was cut into numbered tiles and shuffled. For each original position, name the displayed tile that belongs there.}
{$r\times c$ tiles with number badges; tiers scale grid size ($3\times3$ to $6\times6$).}
{The full piece-order sequence must match exactly.}

\taskcard{rotated jigsaw}
{Tiles are shuffled \emph{and} each rotated by $0/90/180/270^\circ$. For each original position give the displayed tile and the CCW rotation that restores it (e.g., \texttt{3R90}).}
{Square tiles; tiers from $2\times2$ to $3\times3$ with forced rotations at the top tier.}
{The full tile-and-rotation assignment must match exactly.}

\taskcard{restore measure}
{Undo the shift, then report what percentage of the left half is covered by textured polygonal patches ($\pm 1.5$pp).}
{Polygons filled with patches from another photo (no single-color threshold segments them); frame rolled.}
{Percentage within $\pm 1.5$pp.}

\paragraph{Worked examples.} Highest-difficulty instances (validation split, never test) with gold answers.\par\medskip
\begin{tcbitemize}[raster columns=2, colback=white, colframe=black!40, boxrule=0.4pt, arc=0.8mm, left=1.4mm, right=1.4mm, top=1.2mm, bottom=1.2mm, raster equal height=rows, raster column skip=1.6mm, raster row skip=1.6mm]
\tcbitem \includegraphics[width=0.49\linewidth]{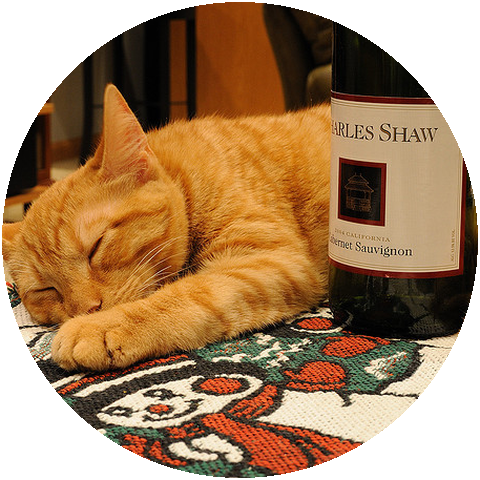}\hfill\includegraphics[width=0.49\linewidth]{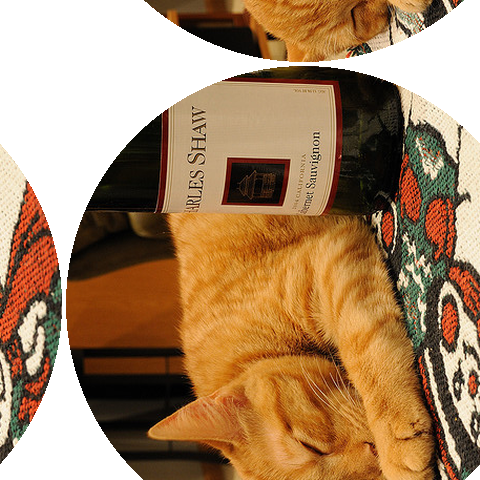}\par\vspace{1pt}
{\scriptsize\textbf{shift align.} What are dx, dy (multiples of 8, between -64 and 64) and theta?\par\vspace{1pt}
\textcolor{tickgreen}{\textbf{Answer:} \texttt{64, 64, 90}}}
\tcbitem \includegraphics[width=0.49\linewidth]{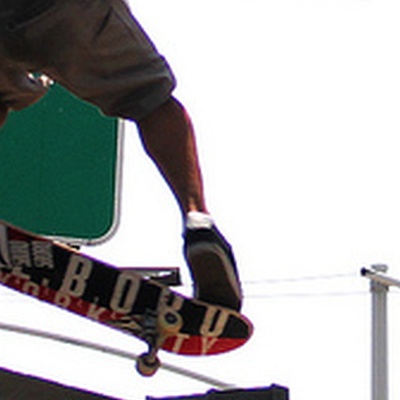}\hfill\includegraphics[width=0.49\linewidth]{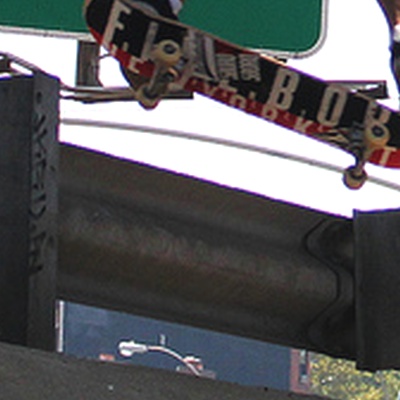}\par\vspace{1pt}
{\scriptsize\textbf{alignment.} What is the content shift (dx, dy) in pixels?\par\vspace{1pt}
\textcolor{tickgreen}{\textbf{Answer:} \texttt{207, -187}}}
\tcbitem \includegraphics[width=0.49\linewidth]{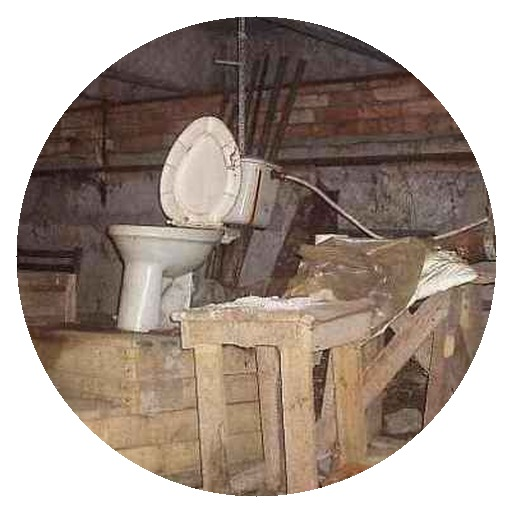}\hfill\includegraphics[width=0.49\linewidth]{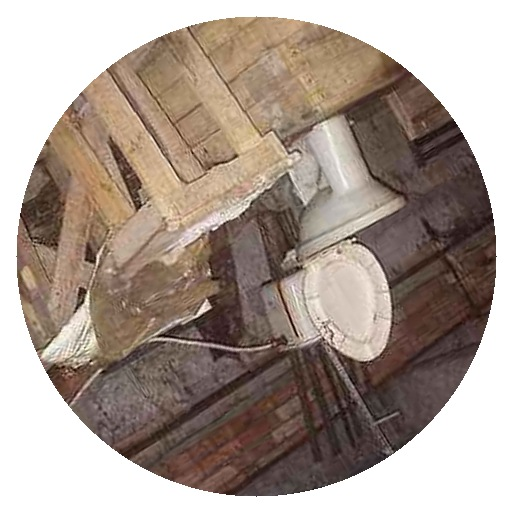}\par\vspace{1pt}
{\scriptsize\textbf{rot scale.} What are theta (in degrees, 0-360) and s?\par\vspace{1pt}
\textcolor{tickgreen}{\textbf{Answer:} \texttt{210.7, 1.069}}}
\tcbitem \includegraphics[width=0.49\linewidth]{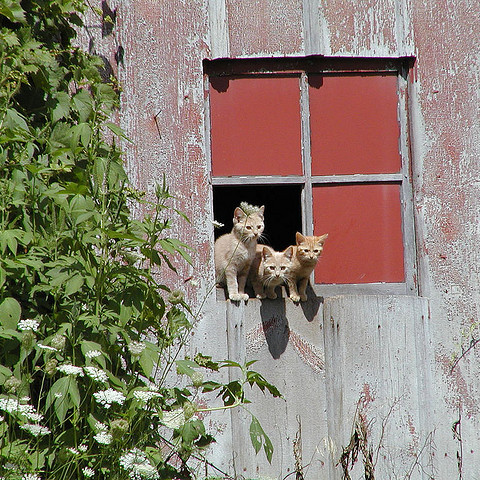}\hfill\includegraphics[width=0.49\linewidth]{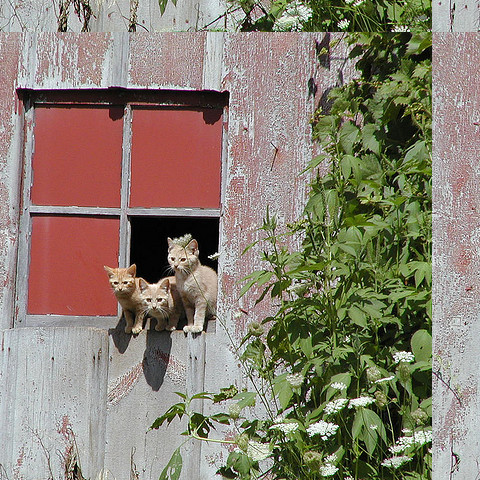}\par\vspace{1pt}
{\scriptsize\textbf{transform chain.} What is the canonical form?\par\vspace{1pt}
\textcolor{tickgreen}{\textbf{Answer:} \texttt{shift(48,32); rot(0); fliph(1)}}}
\end{tcbitemize}
\begin{tcbitemize}[raster columns=4, colback=white, colframe=black!40, boxrule=0.4pt, arc=0.8mm, left=1.4mm, right=1.4mm, top=1.2mm, bottom=1.2mm, raster equal height=rows, raster column skip=1.6mm, raster row skip=1.6mm]
\tcbitem \includegraphics[width=\linewidth]{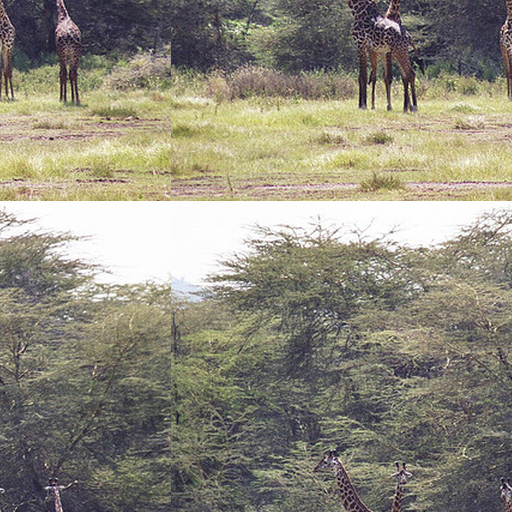}\par\vspace{1pt}
{\scriptsize\textbf{cyclic shift.} Find the shift (dx, dy) in pixels.\par\vspace{1pt}
\textcolor{tickgreen}{\textbf{Answer:} \texttt{171, 201}}}
\tcbitem \includegraphics[width=\linewidth]{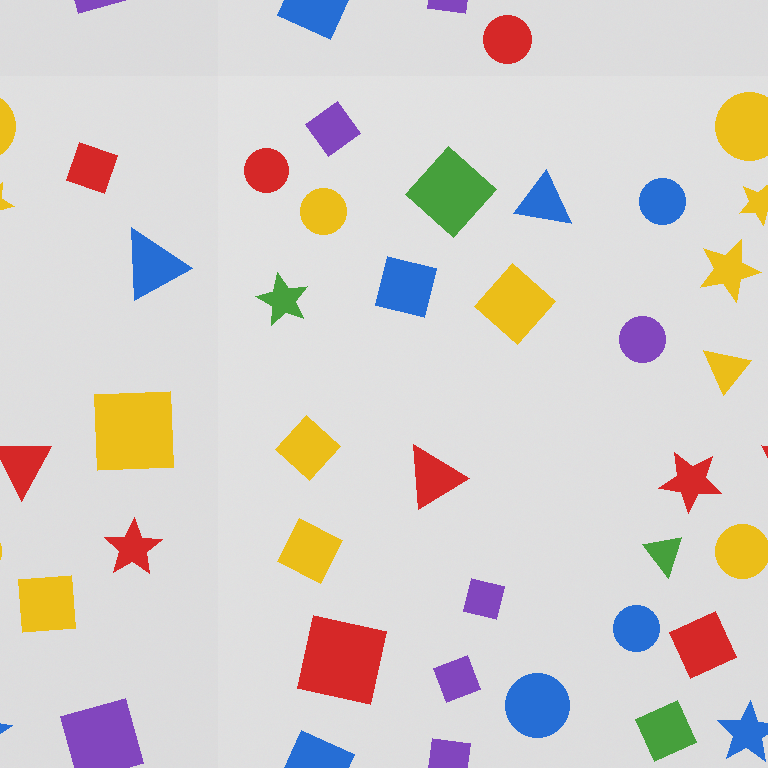}\par\vspace{1pt}
{\scriptsize\textbf{restore count.} Undo the shift to restore the original picture, then count how many shapes lie ENTIRELY in the LEFT HALF of the restored picture.\par\vspace{1pt}
\textcolor{tickgreen}{\textbf{Answer:} \texttt{18}}}
\tcbitem \includegraphics[width=\linewidth]{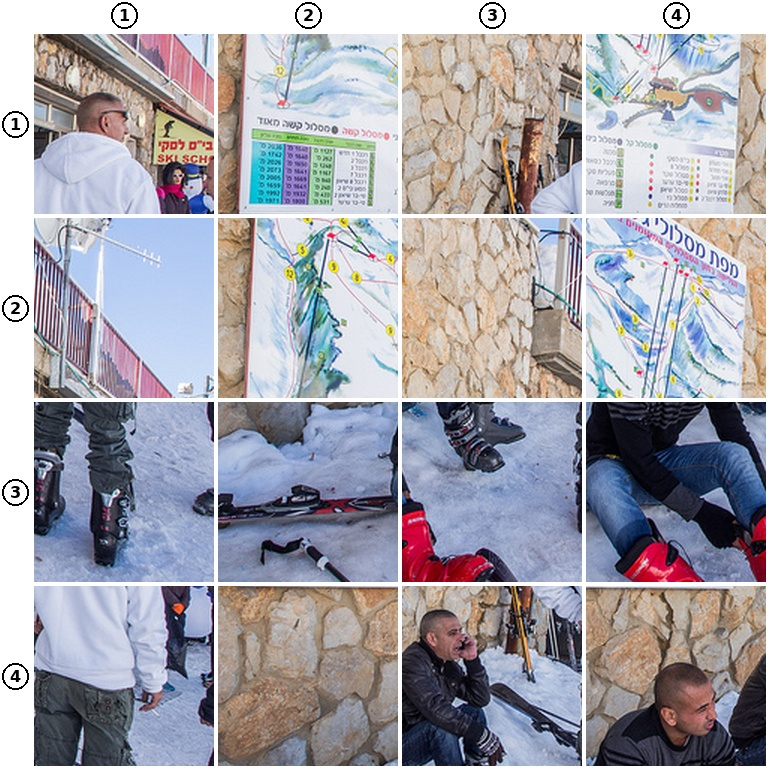}\par\vspace{1pt}
{\scriptsize\textbf{unshuffle 2D.} In what order should the numbered rows be arranged (top to bottom), and the numbered columns (left to right), to reconstruct the original photo?\par\vspace{1pt}
\textcolor{tickgreen}{\textbf{Answer:} \texttt{rows: 2, 1, 4, 3; cols: 2, 4, 3, 1}}}
\tcbitem \includegraphics[width=\linewidth]{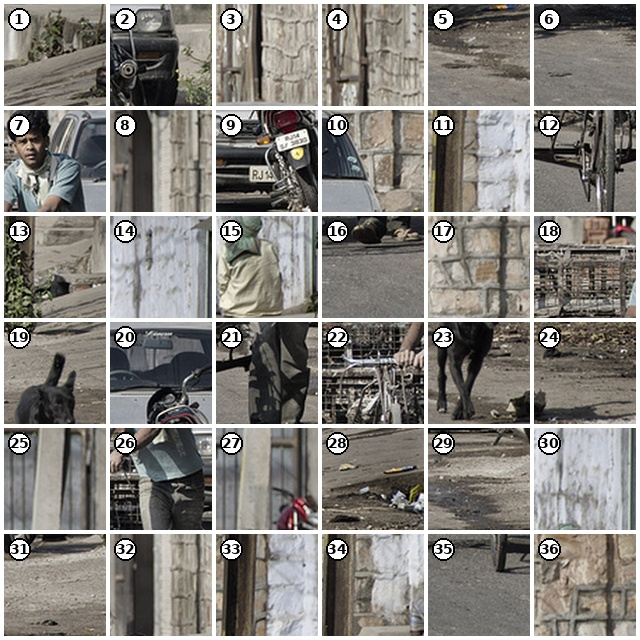}\par\vspace{1pt}
{\scriptsize\textbf{jigsaw.} In what order should the numbered pieces be arranged to reconstruct the original image? Please provide your answer as a sequence indicating where each numbered piece should be placed in the final arrangement.\par\vspace{1pt}
\textcolor{tickgreen}{\textbf{Answer:} \texttt{[25, 8, 4, 36, 33, 14, 27, 32, \dots]\,{\rm(36-piece permutation)}}}}
\tcbitem \includegraphics[width=\linewidth]{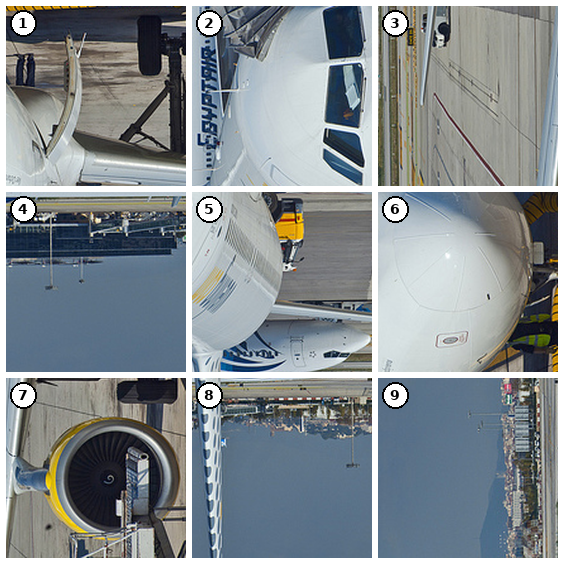}\par\vspace{1pt}
{\scriptsize\textbf{rotated jigsaw.} For each ORIGINAL position, which displayed tile belongs there, and by how many degrees must it be rotated COUNTERCLOCKWISE to restore its original orientation? Answer exactly in this format (one term per original position): 3R90, 1R0, 4R270, 2R0.\par\vspace{1pt}
\textcolor{tickgreen}{\textbf{Answer:} \texttt{6R270, 8R180, 4R180, 5R180, 2R90, 9R270, 3R270, 7R90, 1R90}}}
\tcbitem \includegraphics[width=\linewidth]{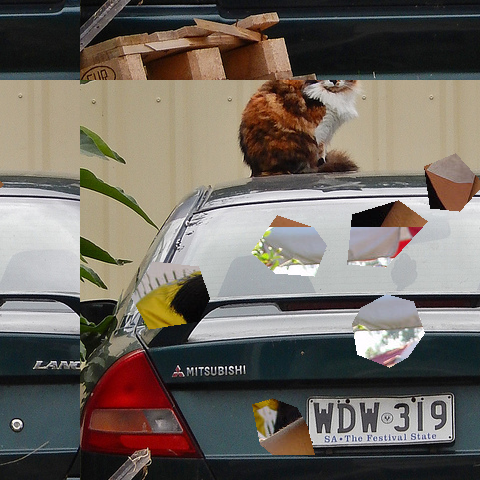}\par\vspace{1pt}
{\scriptsize\textbf{restore measure.} After undoing the shift, what percentage of the LEFT HALF of the original picture's area is covered by the textured polygonal patches?\par\vspace{1pt}
\textcolor{tickgreen}{\textbf{Answer:} \texttt{8.3}}}
\tcbitem \includegraphics[width=\linewidth]{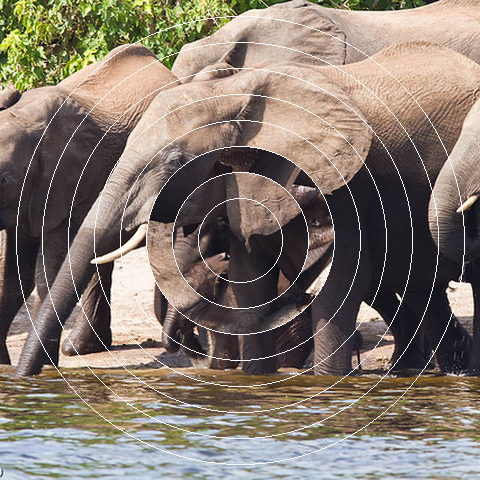}\par\vspace{1pt}
{\scriptsize\textbf{ring rotor.} Concentric rings; exactly one was rotated. Which ring, and by how many degrees?\par\vspace{1pt}
\textcolor{tickgreen}{\textbf{Answer:} \texttt{2, 140.4}}}
\tcbitem \includegraphics[width=\linewidth]{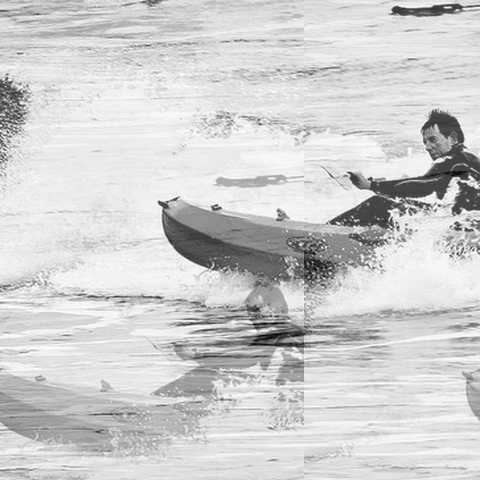}\par\vspace{1pt}
{\scriptsize\textbf{double exposure.} The image blends a dominant photo with fainter cyclically shifted ghosts. Report the shift (dx, dy) of every ghost, heavier first.\par\vspace{1pt}
\textcolor{tickgreen}{\textbf{Answer:} \texttt{-176, 171}}}
\end{tcbitemize}

\taskcard{ring rotor}
{An image is divided into concentric rings and exactly one ring is rotated. Report the ring index and signed angle (index exact, angle within $\pm5^\circ$).}
{A COCO photograph is remapped into annuli with visible boundaries; one annulus receives an item-specific rotation and tiers vary ring count and ambiguity.}
{Ring index exact and angle within $\pm 5^\circ$.}

\taskcard{double exposure}
{An image mixes a dominant photograph with one or two fainter cyclically shifted copies. Recover every signed shift, ordered from the heavier to lighter ghost (each component within $\pm8$px).}
{One or two wrap-around translations with distinct weights are composited with the source; tiers vary ghost count, strength, and separation.}
{Every signed shift within $\pm 8$px; ghost count and order must match.}

\subsection{Perception \& counting}
\label{app:tasks:percep}

\taskcard{grating period}
{A faint sinusoidal stripe grating was added over the whole photo. Report the stripe period in pixels (one decimal, $\pm 1.5$px).}
{A low-amplitude sinusoid at a random orientation and period is added to a COCO photo; tiers lower the amplitude and widen the period range.}
{Period within $\pm 1.5$px.}

\taskcard{anomaly grid}
{A photo is divided into a $6\times6$ grid whose cells all contain the same kind of object with small natural variations; a few cells are anomalous. List all anomalous cells.}
{$K$ cells receive a locally different object; an equal-strength global brightness/noise perturbation is applied to the whole image.}
{The reported cell set must equal the true anomaly set.}

\taskcard{peak count}
{The dark image contains faint glowing spots in background noise. How many distinct spots are there? Answer: one integer (exact).}
{$N$ soft Gaussian blobs at low signal-to-noise are placed over noise; tiers raise $N$ and lower the SNR.}
{The reported integer must equal the true count.}

\taskcard{congruent pair}
{Several outlined shapes are shown; exactly one pair is congruent up to rotation/flip. Name the pair.}
{Shapes generated with controlled pairwise dissimilarity margins; one planted congruent pair.}
{The reported index pair must equal the true congruent pair.}

\taskcard{trace length}
{A single continuous winding, self-crossing cyan path is drawn. Report its length in pixels divided by $200$ (one decimal, $\pm 0.3$).}
{A random winding polyline of controlled total length is rendered at fixed width; tiers add length, sharper bends, and self-crossings.}
{Length$/200$ within $\pm 0.3$.}

\paragraph{Worked examples.} Highest-difficulty instances (validation split, never test) with gold answers.\par\medskip
\begin{tcbitemize}[raster columns=4, colback=white, colframe=black!40, boxrule=0.4pt, arc=0.8mm, left=1.4mm, right=1.4mm, top=1.2mm, bottom=1.2mm, raster equal height=rows, raster column skip=1.6mm, raster row skip=1.6mm]
\tcbitem \includegraphics[width=\linewidth]{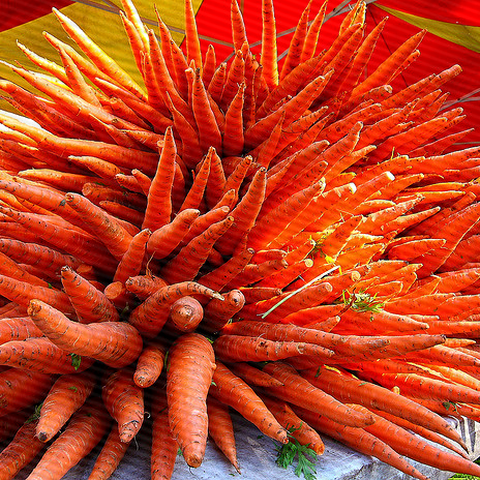}\par\vspace{1pt}
{\scriptsize\textbf{grating period.} What is the period of the stripes, in pixels from one stripe to the next?\par\vspace{1pt}
\textcolor{tickgreen}{\textbf{Answer:} \texttt{8.2}}}
\tcbitem \includegraphics[width=\linewidth]{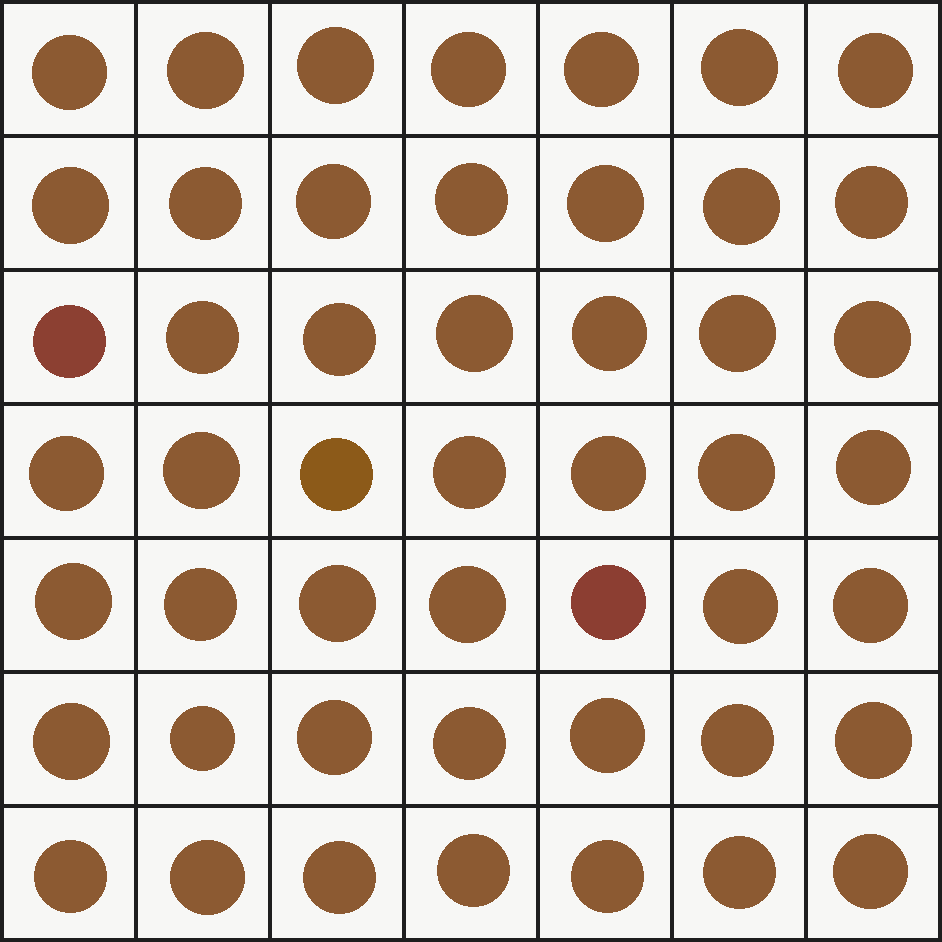}\par\vspace{1pt}
{\scriptsize\textbf{anomaly grid.} A $7\times 7$ grid of similar objects; a few cells are anomalous beyond natural variation. Identify every anomalous cell.\par\vspace{1pt}
\textcolor{tickgreen}{\textbf{Answer:} \texttt{15, 24, 33, 37}}}
\tcbitem \includegraphics[width=\linewidth]{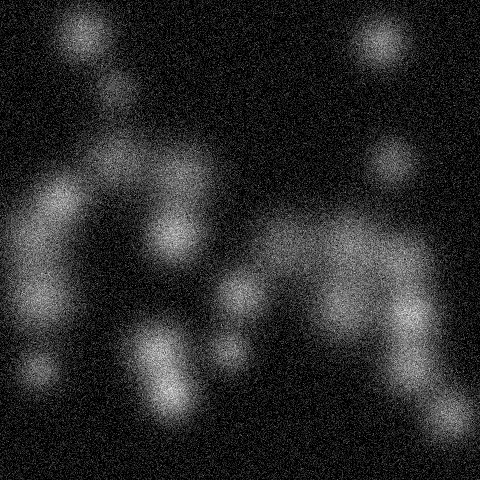}\par\vspace{1pt}
{\scriptsize\textbf{peak count.} How many distinct glowing spots are there?\par\vspace{1pt}
\textcolor{tickgreen}{\textbf{Answer:} \texttt{22}}}
\tcbitem \includegraphics[width=\linewidth]{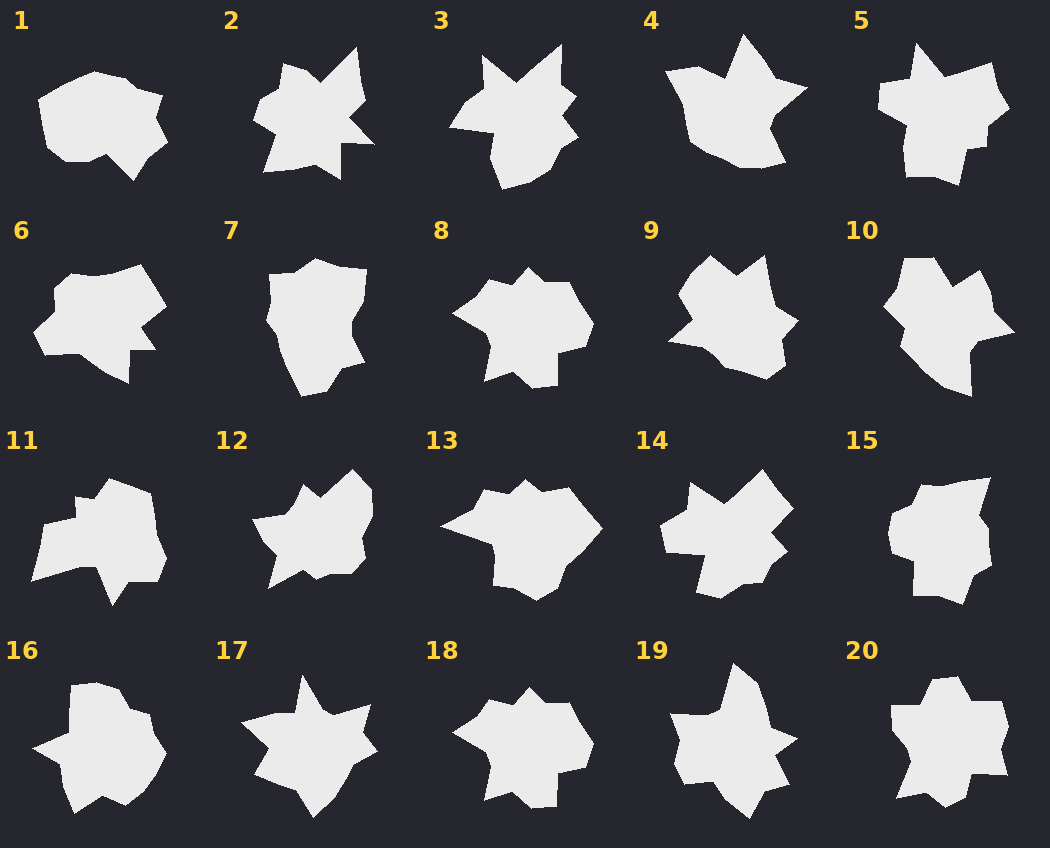}\par\vspace{1pt}
{\scriptsize\textbf{congruent pair.} Which two shapes are congruent?\par\vspace{1pt}
\textcolor{tickgreen}{\textbf{Answer:} \texttt{4, 7}}}
\tcbitem \includegraphics[width=\linewidth]{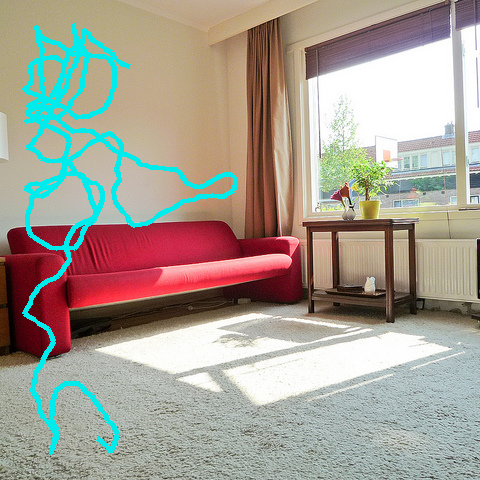}\par\vspace{1pt}
{\scriptsize\textbf{trace length.} How long is the cyan path in pixels? Report the length in pixels DIVIDED BY 200, to one decimal (for example, a 730-pixel path is 3.7).\par\vspace{1pt}
\textcolor{tickgreen}{\textbf{Answer:} \texttt{12.3}}}
\tcbitem \includegraphics[width=\linewidth]{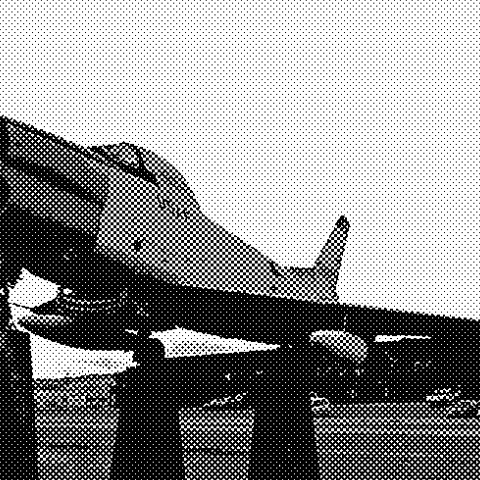}\par\vspace{1pt}
{\scriptsize\textbf{scan moir\'e.} A halftone print was rescanned at an unknown scale $r$. Report $r$ to within $\pm 0.004$.\par\vspace{1pt}
\textcolor{tickgreen}{\textbf{Answer:} \texttt{0.9811}}}
\end{tcbitemize}

\taskcard{scan moir\'e}
{A halftone print with known dot pitch is re-scanned at an unknown scale. Recover the scale factor from the resulting moir\'e pattern (within $\pm0.004$).}
{A COCO photograph is halftoned on a $6$px lattice, resized by a non-unit scale, and center-cropped; tiers move the spectral peak progressively closer to the nominal frequency.}
{Scale $r$ within $\pm 0.004$.}

\subsection{Localization}
\label{app:tasks:local}

\taskcard{crop origin}
{The second image is a small detail cropped from the first (then slightly resized and re-compressed). Locate the source rectangle in the first image (box, IoU$\ge 0.5$; coordinates normalized to $0$--$1000$).}
{A random sub-rectangle is cropped, resized and re-JPEG'd; tiers shrink the crop and add near-duplicate decoy regions.}
{Box IoU $\ge 0.5$; coordinates normalized $0$--$1000$.}

\taskcard{brightest blob}
{Among several glowing blobs of differing intensity, report the center of the single brightest one (normalized $0$--$1000$, $\pm 2$px).}
{Multiple Gaussian blobs of differing peak intensity are placed over noise; tiers narrow the intensity gap between the brightest and the runner-up.}
{Centre point within $\pm 2$px (on the $480$px canvas).}

\taskcard{rect swap}
{Two same-size rectangles in image 2 had their contents swapped. Report both rectangles as \texttt{x1,y1,x2,y2} normalized to $0$--$1000$, order-free.}
{Rectangles snapped to an 8px grid; contents exchanged.}
{Both boxes IoU $\ge 0.5$, order-free; coordinates normalized $0$--$1000$.}

\taskcard{retouch region}
{One rectangular region was texture-smoothed (a blur/beauty filter) with soft edges. Locate the retouched rectangle (box, IoU$\ge 0.5$; normalized $0$--$1000$).}
{A random rectangle is low-pass filtered and feathered into a COCO photo; tiers soften the boundary and shrink the region.}
{Box IoU $\ge 0.5$; coordinates normalized $0$--$1000$.}

\taskcard{circle fit}
{Most dots lie along a circular arc (with scatter); a few are distractors. If the arc were completed into a full circle, report its center (normalized $0$--$1000$, $\pm 4$px).}
{Dots are sampled along an arc with Gaussian scatter plus random off-arc distractors; tiers raise the scatter and shorten the arc.}
{Constructive (circle known); an identifiability screen requires the arc-inlier fit to be unique and stable against distractor-only fits.}

\paragraph{Worked examples.} Highest-difficulty instances (validation split, never test) with gold answers.\par\medskip
\begin{tcbitemize}[raster columns=2, colback=white, colframe=black!40, boxrule=0.4pt, arc=0.8mm, left=1.4mm, right=1.4mm, top=1.2mm, bottom=1.2mm, raster equal height=rows, raster column skip=1.6mm, raster row skip=1.6mm]
\tcbitem \includegraphics[width=0.49\linewidth]{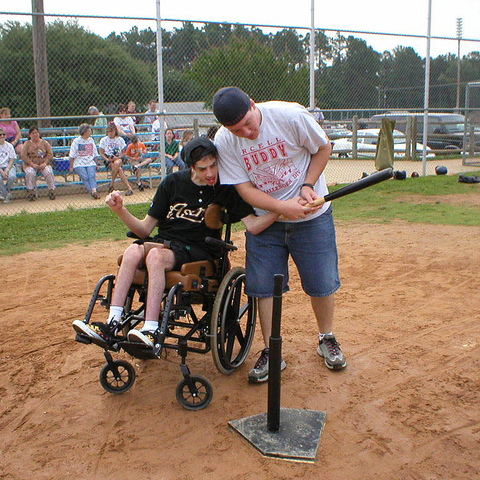}\hfill\includegraphics[width=0.49\linewidth]{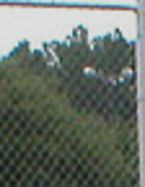}\par\vspace{1pt}
{\scriptsize\textbf{crop origin.} Where in the first image was the detail taken from? Report the crop rectangle in first-image coordinates as x1,y1,x2,y2 (normalized 0--1000).\par\vspace{1pt}
\textcolor{tickgreen}{\textbf{Answer:} \texttt{767,600,917,733}}}
\tcbitem \includegraphics[width=0.49\linewidth]{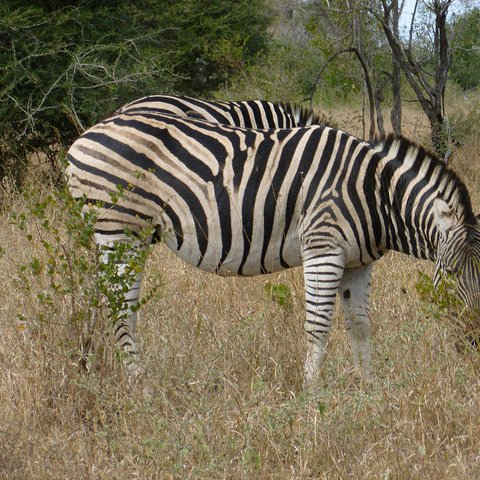}\hfill\includegraphics[width=0.49\linewidth]{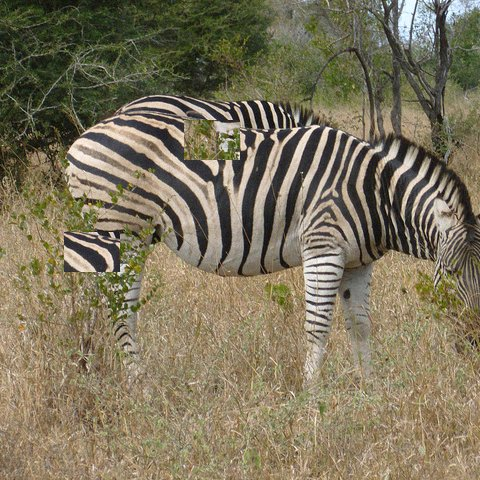}\par\vspace{1pt}
{\scriptsize\textbf{rect swap.} Which two rectangles were swapped? Give each as x1,y1,x2,y2 (normalized 0--1000).\par\vspace{1pt}
\textcolor{tickgreen}{\textbf{Answer:} \texttt{383,250,500,333 $\leftrightarrow$ 133,483,250,567}}}
\end{tcbitemize}
\begin{tcbitemize}[raster columns=4, colback=white, colframe=black!40, boxrule=0.4pt, arc=0.8mm, left=1.4mm, right=1.4mm, top=1.2mm, bottom=1.2mm, raster equal height=rows, raster column skip=1.6mm, raster row skip=1.6mm]
\tcbitem \includegraphics[width=\linewidth]{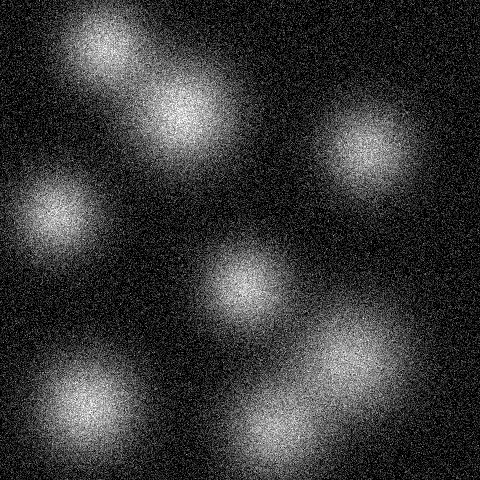}\par\vspace{1pt}
{\scriptsize\textbf{brightest blob.} Give the coordinates x,y of the CENTER of the single brightest blob, (normalized 0--1000).\par\vspace{1pt}
\textcolor{tickgreen}{\textbf{Answer:} \texttt{379,235}}}
\tcbitem \includegraphics[width=\linewidth]{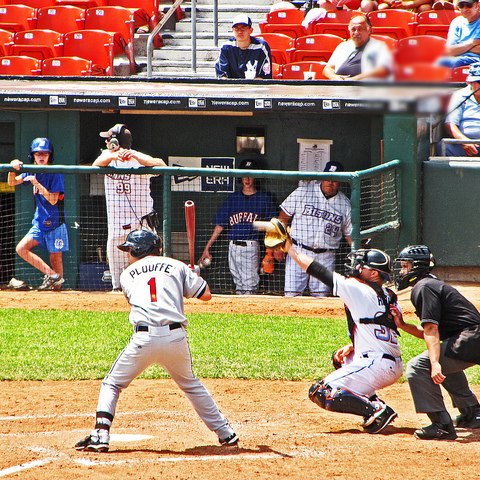}\par\vspace{1pt}
{\scriptsize\textbf{retouch region.} Locate the retouched rectangle. Report it as x1,y1,x2,y2 (normalized 0--1000).\par\vspace{1pt}
\textcolor{tickgreen}{\textbf{Answer:} \texttt{750,83,950,250}}}
\tcbitem \includegraphics[width=\linewidth]{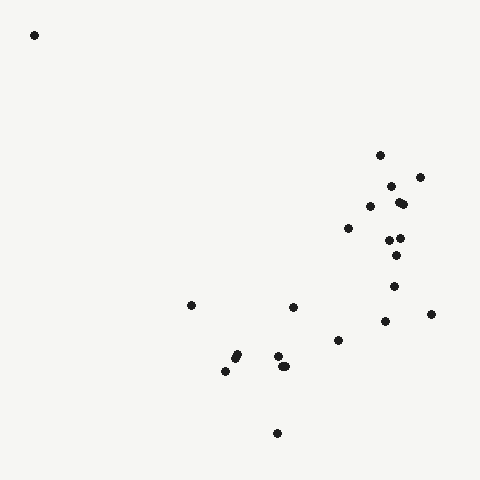}\par\vspace{1pt}
{\scriptsize\textbf{circle fit.} If the arc were completed into a full circle, where would its CENTER be? Report the center coordinates x,y, (normalized 0--1000).\par\vspace{1pt}
\textcolor{tickgreen}{\textbf{Answer:} \texttt{552,479}}}
\tcbitem \includegraphics[width=\linewidth]{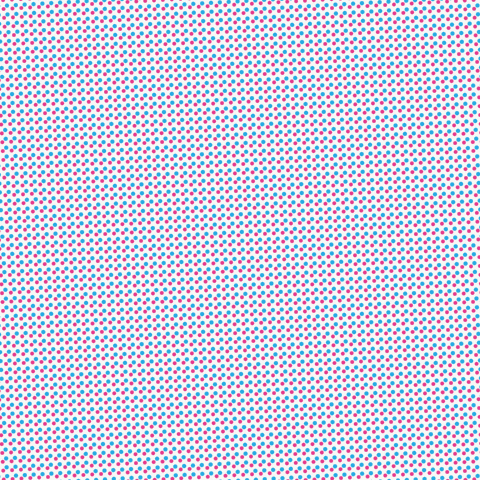}\par\vspace{1pt}
{\scriptsize\textbf{print misreg.} A cyan and a magenta halftone lattice share pitch and rotation, but one layer is slightly offset. Report the offset (dx, dy) in pixels.\par\vspace{1pt}
\textcolor{tickgreen}{\textbf{Answer:} \texttt{-3.39, 3.39}}}
\tcbitem \includegraphics[width=\linewidth]{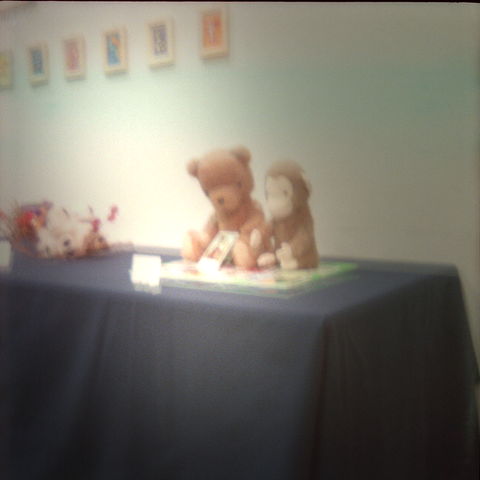}\par\vspace{1pt}
{\scriptsize\textbf{gain center.} Report the centers of all spotlights, ordered from strongest to weakest peak gain.\par\vspace{1pt}
\textcolor{tickgreen}{\textbf{Answer:} \texttt{195, 319; 117, 118}}}
\end{tcbitemize}

\taskcard{print misreg}
{Two rotated colour halftone lattices share pitch and orientation, but one is shifted by a signed sub-pixel offset. Recover $(dx,dy)$ within $\pm0.3$px.}
{Cyan and magenta dot layers are rendered with item-specific pitch and rotation, then displaced by a non-integer offset smaller than half a pitch.}
{$(dx,dy)$ each within $\pm 0.3$px.}

\taskcard{gain center}
{Given a reference image and a relit copy with two or three soft spotlights plus an ambient tilt, report every spotlight centre ordered by strength (within $\pm15$px).}
{Pairwise-distinct Gaussian gain fields and a linear illumination plane are multiplied into the source photograph; tiers vary the number, strength, width, and overlap of lights.}
{Every spotlight centre within $\pm 15$px, in the given strength order.}

\subsection{Color}
\label{app:tasks:color}

\taskcard{film color}
{A rectangular paint swatch is partly covered by a $70\%$-opacity translucent colour film. Report the swatch's true underlying RGB (per-channel $\pm 10$).}
{A flat swatch is painted, then a translucent coloured strip is alpha-composited across its middle; the top and bottom of the swatch stay uncovered.}
{The reported RGB must be within $\pm 10$ on each channel.}

\taskcard{swatch RGB}
{A black arrow points at exactly one of several solid-colour swatches. Report that swatch's RGB (per-channel $\pm 10$).}
{Flat, noise-free swatches are painted on the photo; an outlined arrow's tip touches the target swatch; distractor line segments may be added.}
{The reported RGB must be within $\pm 10$ on each channel.}

\taskcard{patch means}
{A red rectangle marks a region; magenta annotation dots were sprinkled inside it after capture. Report the average RGB of the underlying photo inside the rectangle, ignoring the dots (per-channel $\pm 10$).}
{The rectangle is drawn on a COCO photo and annotation dots are softly composited inside; the target is the mean of the original pixels under the rectangle.}
{The reported RGB must be within $\pm 10$ on each channel.}

\taskcard{color count}
{Several solid-colour stickers are pasted; some share the exact same fill and some differ only subtly. How many distinct fill colours are there? Answer: one integer (exact).}
{Stickers with a controlled number of distinct fills plus $\pm3$/channel noise; tiers add near-duplicate colour pairs.}
{The reported integer must equal the true count.}

\paragraph{Worked examples.} Highest-difficulty instances (validation split, never test) with gold answers.\par\medskip
\begin{tcbitemize}[raster columns=4, colback=white, colframe=black!40, boxrule=0.4pt, arc=0.8mm, left=1.4mm, right=1.4mm, top=1.2mm, bottom=1.2mm, raster equal height=rows, raster column skip=1.6mm, raster row skip=1.6mm]
\tcbitem \includegraphics[width=\linewidth]{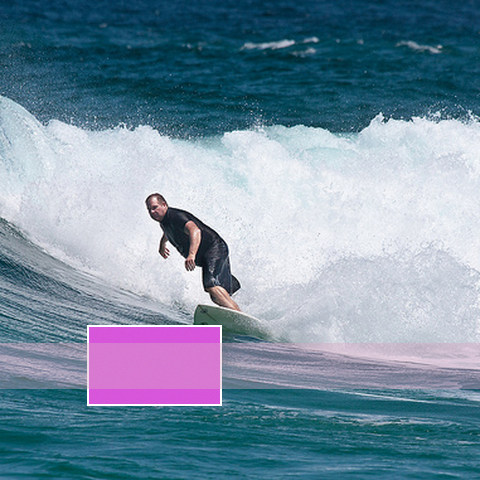}\par\vspace{1pt}
{\scriptsize\textbf{film color.} What is the colour of the FILM itself?\par\vspace{1pt}
\textcolor{tickgreen}{\textbf{Answer:} \texttt{80,73,227}}}
\tcbitem \includegraphics[width=\linewidth]{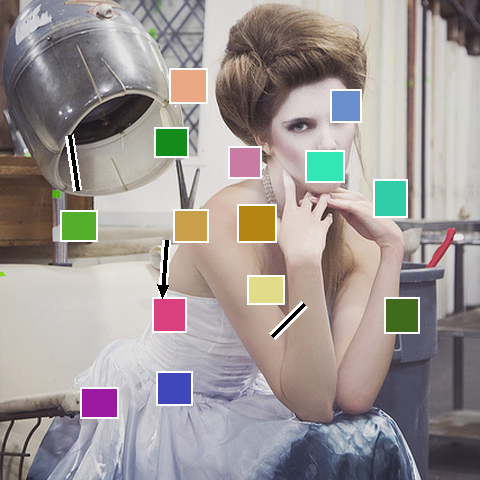}\par\vspace{1pt}
{\scriptsize\textbf{swatch RGB.} What is the exact fill colour (RGB) of the swatch the arrow points to?\par\vspace{1pt}
\textcolor{tickgreen}{\textbf{Answer:} \texttt{168,199,159}}}
\tcbitem \includegraphics[width=\linewidth]{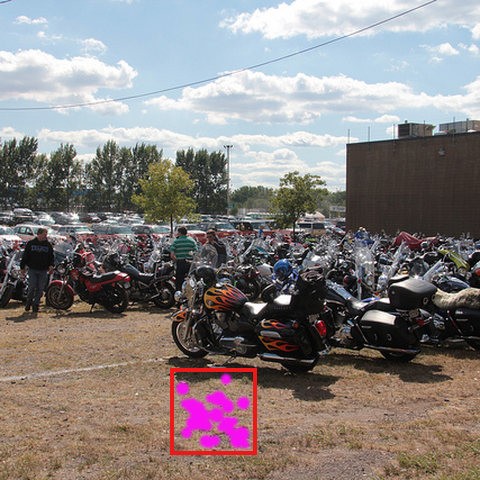}\par\vspace{1pt}
{\scriptsize\textbf{patch means.} What is the AVERAGE colour (mean RGB) of the PHOTO pixels strictly inside the rectangle? Exclude the red border and everything the magenta dots cover.\par\vspace{1pt}
\textcolor{tickgreen}{\textbf{Answer:} \texttt{134,111,92}}}
\tcbitem \includegraphics[width=\linewidth]{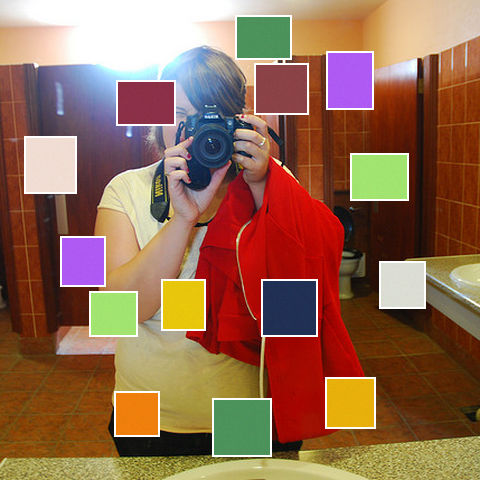}\par\vspace{1pt}
{\scriptsize\textbf{color count.} How many DISTINCT fill colours are used across all stickers? Two stickers count as the same colour only if their underlying fill is identical; a subtle-but-real difference counts as a different colour.\par\vspace{1pt}
\textcolor{tickgreen}{\textbf{Answer:} \texttt{11}}}
\tcbitem \includegraphics[width=\linewidth]{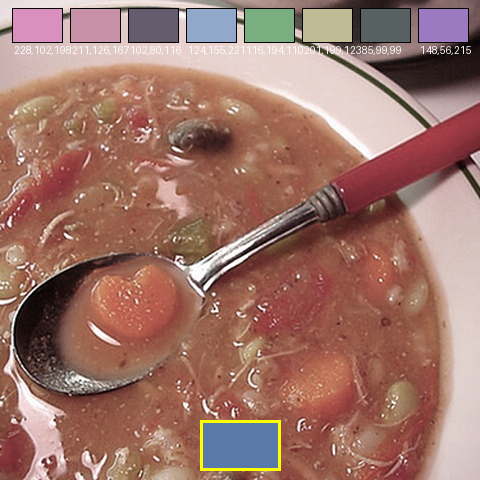}\par\vspace{1pt}
{\scriptsize\textbf{channel mix.} The image was pushed through an unknown channel-mixing matrix. Recover the target patch's true sRGB colour (each channel within $\pm 8$).\par\vspace{1pt}
\textcolor{tickgreen}{\textbf{Answer:} \texttt{32, 98, 191}}}
\end{tcbitemize}

\taskcard{channel mix}
{A camera applies unknown $3\times3$ channel crosstalk. Using eight labelled calibration patches, recover the true sRGB colour of a held-out target patch (per-channel $\pm8$).}
{A near-identity mixing matrix acts in linear-light space before conversion back to sRGB; tiers increase off-diagonal crosstalk.}
{The reported RGB must be within $\pm 8$ on each channel.}

\section{Reflection prompts (verbatim)}
\label{app:prompts}
Both prompts below are task-agnostic: the identical string is used for every task; the task never appears by name. The observational prompt is the single-call reflection; the \methodshort{} prompt drives the sandboxed debugging session (rendered images from executed code are returned to the model each step).

\begin{promptbox}{Prompt for Observational Reflection}\small\ttfamily
You previously attempted several instances of ONE task type using a stateful Python sandbox (the transcripts, your code, its outputs\{img\_note\}, the ground-truth answers, and your wrong answers are shown above).\par\medskip
Reflect on WHY your attempts failed, then EDIT your skill library for this task type. Your future self will be given this library before attempting new instances.\par\medskip
Each skill MUST have exactly three sections:\\
When: <trigger condition>\\
Procedure: <numbered concrete steps>\\
Verify: <which INTERMEDIATE quantity to RENDER as an image (not just print), what it should look like when the procedure is working, and what it looks like when something is silently wrong>\par\medskip
Rules:\\
- Express changes as explicit edit operations:\\
  \mbox{<edit op="add" name="short-skill-name">}When: ...\\
Procedure:\\
1. ...\\
Verify: ...</edit>\\
  <edit op="replace" name="existing-name">...new full body...</edit>\\
  <edit op="delete" name="existing-name"></edit>\\
- At most 3 edit operations per reflection. Keep the library at <= 6 skills, each under 250 words (the Verify section does not count toward the limit).\\
- General: must work for ANY instance of this task type. Never include answers, coordinates or values specific to the examples above.\\
- If the same skill has been edited repeatedly without improvement (see the attendance table), question the METHOD itself rather than its parameters -- consider an alternative approach as a new skill.\\
- Check the Verify sections: is there any intermediate computation your procedures rely on that is never rendered and inspected? Silent numeric bugs are invisible in printed numbers but obvious in a rendered image.\\
- Some attempts shown above SUCCEEDED: identify what worked there and do not break it.\\
- If the remaining failures look inherently hard and the library already covers them as well as it can, you may output exactly NO-EDIT instead of forcing changes.\par\medskip
Output ONLY the edit operations (or NO-EDIT).
\end{promptbox}

\begin{promptbox}{Prompt for Executable Reflection}\small\ttfamily
You are debugging your own past failures on ONE task type, in a live Python sandbox.\par\medskip
Preloaded variables (already in the sandbox):\\
- cases: list of dicts, one per case: cases[i]['images'] (list of PIL images), cases[i]['gold'] (ground-truth answer string), cases[i]['pred'] (your wrong answer; None if this was a SUCCESSFUL case), cases[i]['code'] (list of the code strings you ran in that attempt), cases[i]['prompt'] (the task instruction)\\
- Successful cases (pred is None) show approaches of yours that WORKED.\par\medskip
Your job, in this order:\\
1. INVESTIGATE with code: re-run suspect computations from your failing code on the actual images, print intermediate values AND dtypes, render intermediate arrays as images and look at them, and test hypotheses by checking against the known gold. Compare, line by line, how your successful and failing attempts compute the same quantity. Find the ROOT CAUSE, not the symptom.\\
2. Verify the fix: show with code that the corrected computation recovers the gold on at least 2 of the failing cases.\\
3. Only then, output skill-library edits in EXACTLY this format (and nothing after them):\\
\mbox{<edit op="add" name="short-skill-name">}When: ...\\
Procedure:\\
1. ...\\
Verify: ...</edit>\\
(ops: add / replace / delete; at most 3 edits; library <= 6 skills; each skill <= 250 words; every skill must have When/Procedure/Verify; Verify = which intermediate to RENDER and what correct vs broken looks like. Skills must be GENERAL for this task type — never include answers or values specific to these cases.)\par\medskip
Write ```python code blocks to use the sandbox; I will run them and return outputs. When you are confident, emit the edits.
\end{promptbox}

\begin{promptbox}{Forced Wrap-Up (issued when the investigation budget is exhausted)}\small\ttfamily
Investigation budget is exhausted. Based on everything you verified above, output your skill-library edit operations NOW, in exactly the format specified (at most 3 edit ops; each skill When/Procedure/Verify; general, no case-specific values). Output ONLY the edit operations.
\end{promptbox}

\section{Exhibits: a delivered skill and a reflection session}
\label{app:exhibits}
\subsection{A delivered multimodal skill, verbatim}
\label{app:exhibits:skill}
The complete text of the delivered library --- one skill, 208 words, written entirely by the model. At solve time this text is appended to the system prompt and the skill's two anchor images (harvested from the model's own successful runs) are placed before the task images.
\begin{promptbox}{Delivered Skill: robust-shift-estimation (alignment, v7)}
\begin{Verbatim}[breaklines,fontsize=\small]
### Skill: robust-shift-estimation
When: Task requires calculating pixel shift (dx, dy) between two overlapping crops of the same image.
Procedure:
1. Convert both input images to grayscale using OpenCV (`cv2.cvtColor` with `COLOR_RGB2GRAY`).
2. Detect keypoints and compute descriptors in both images using SIFT (`cv2.SIFT_create()`) if available; otherwise use ORB (`cv2.ORB_create(nfeatures=500)`). SIFT is preferred for robustness against scale and rotation changes often present in crops.
3. Match descriptors using `cv2.BFMatcher`. For SIFT use `NORM_L2`, for ORB use `NORM_HAMMING`. Use `knnMatch(k=2)`.
4. Apply Lowe's ratio test: keep match m if `m.distance < 0.75 * n.distance` where n is the second-best match.
5. If fewer than 4 matches pass the ratio test, fall back to `crossCheck=True` matching.
6. Use RANSAC (`cv2.findHomography` with `cv2.RANSAC`) to filter outliers from the good matches. Ensure at least 4 inliers remain.
7. Calculate coordinate differences for each RANSAC-filtered match: `dx = pt2.x - pt1.x`, `dy = pt2.y - pt1.y`.
8. Compute the **median** of all `dx` values and all `dy` values to determine the final shift. Round to nearest integer.
9. Render a visualization: Create copies of original images. Draw red circles on Image 1 at matched keypoint locations from Image 1. Draw green circles on Image 2 at matched keypoint locations from Image 2. Display them side-by-side.
Verify: The rendered image must show red dots on Image 1 and green dots on Image 2 highlighting identical visual features (e.g., specific corners, texture patches that correspond between images). If the procedure fails, the dots will appear on unrelated textures or random noise, indicating incorrect feature matching. This visual check is critical before trusting the calculated median shift. Additionally, print the number of matches found after each filtering step (ratio test, fallback, RANSAC) to ensure the pipeline is not silently failing due to lack of matches.
\end{Verbatim}
\end{promptbox}
\subsection{Executable-reflection session excerpts}
\label{app:exhibits:session}
Machine-extracted excerpts from a sandbox debugging session on \texttt{cyclic shift}, launched after observational reflection had stalled (full transcript in the artifact release).
\begin{promptbox}{Executable-Reflection Session (cyclic shift), machine-extracted excerpts}
\begin{Verbatim}[breaklines,breakanywhere,fontsize=\footnotesize]
ASSISTANT (excerpt): (round 0) ===== I'll start by investigating the failing cases to understand what went wrong. Let me first examine the structure of the data and compare successful vs failed attempts. ```python # Examine the structure of cases and compare successful vs failed attempts print(f"Total cases: {len(cases)}") for i, case in enumerate(cases): print(f"\nCase {i}: gold={case['gold']}, pred={case['pred']}") if case['images']: img = case['images'][0] print(f" Image shape: {img.size}") if case['code']: print(f" Code length: {len(case['code'])} lines")

EXEC OUTPUT (excerpt):  ===== Total cases: 8 Case 0: gold=370, 413, pred=312, 348 Image shape: (512, 512) Code length: 2 lines Line 0: import numpy as np import matplotlib.pyplot as plt # Load the image img = image... Line 1: ... Case 1: gold=192, 290, pred=325, 203 Image shape: (512, 512) Code length: 3 lines Line 0: import numpy as np from PIL import Image # Load the image img = image_clue[0] a... Line 1:

ASSISTANT (excerpt): (round 1) ===== Let me look at the actual code from successful vs failed cases to understand what's different: ```python # Examine the full code from successful and failed cases print("=" * 80) print("SUCCESSFUL CASES:") print("=" * 80) for i in [6, 7]: # Successful cases print(f"\n--- Case {i} (gold={cases[i]['gold']}) ---") for j, code_line in enumerate(cases[i]['code']): print(f"Code block {j}:") print(code_line[:500]) print("...") print("\n" + "=" * 80) print("FAILED CASES:") print("=" * 80) for i in range(6): # Failed cases print(f"\n--- C

EXEC OUTPUT (excerpt):  ===== ================================================================================ SUCCESSFUL CASES: ================================================================================ --- Case 6 (gold=370, 413) --- Code block 0: import numpy as np from PIL import Image # Load the image img = image_clue[0] arr = np.array(img) h, w, _ = arr.shape print(f"Image shape: {h}x{w}") # The seams should be where ther

ASSISTANT (excerpt): (round 2) ===== Let me see the full code from both successful and failed cases to understand the difference: ```python # Get full code strings for case_idx in [6, 7]: # Successful print(f"\n{'='*80}") print(f"SUCCESSFUL CASE {case_idx}: gold={cases[case_idx]['gold']}") print('='*80) for i, code in enumerate(cases[case_idx]['code']): print(f"\n--- Code block {i} ---") print(code) print("\n\n" + "="*80) print("FAILED CASES:") print("="*80) for case_idx in range(6): # Failed print(f"\n{'='*80}") print(f"FAILED CASE {case_idx}: gold={cases[case
\end{Verbatim}
\end{promptbox}

\section{Full pseudocode}
\label{app:algo}
Algorithms~\ref{alg:loop-full} and~\ref{alg:evr} give the unabridged versions of the loop summarized in Algorithm~\ref{alg:loop}, including the phase structure, incumbent bookkeeping, and the forced wrap-up of the interventional phase.

\begin{algorithm}[H]
\caption{\textsc{SkillLearning} (detailed): one task, frozen model (one restart)}
\label{alg:loop-full}
\begin{algorithmic}[1]
\STATE \COMMENT{\textbf{Phase 0: collect bare \cwi{} experience (no skill)}}
\STATE solve the full training slice with the standard protocol; keep every trajectory (code, outputs, rendered images, answer)
\STATE library $\leftarrow$ empty;\; incumbent $\leftarrow$ none
\medskip
\STATE \COMMENT{\textbf{Phase 1: evolve the skill library} (fixed round budget with early stopping)}
\FOR{round $= 1, 2, \dots$}
   \STATE cases $\leftarrow$ stratified failures (one per difficulty tier) $+$ organic successes from Phase~0
   \IF{the probe score has stalled}
      \STATE edits $\leftarrow$ \textsc{ExecutableReflect}(cases, library) \COMMENT{Algorithm~\ref{alg:evr}: experiment, then edit}
   \ELSE
      \STATE edits $\leftarrow$ one reflection call that reads the cases (with images) and proposes edits
   \ENDIF
   \STATE apply the bounded edit ops to library \COMMENT{skills inside the incumbent cannot be deleted}
   \STATE probeScore $\leftarrow$ accuracy of library on the fixed probe items
   \STATE remember (library, probeScore); continue even if the score dropped
\ENDFOR
\medskip
\STATE \COMMENT{\textbf{Phase 2: select and gate} (on held-out validation items reflection has never seen)}
\STATE evaluate the incumbent on the validation set \COMMENT{if it fails the gate, the runner-up is tested once}
\IF{validation score $\geq$ bare \cwi{} $+$ margin (\eqref{eq:gate})}
   \STATE \textbf{deliver} that library
\ELSE
   \STATE \textbf{deliver nothing} \COMMENT{retain bare \cwi{} for this task}
\ENDIF
\end{algorithmic}
\end{algorithm}

\begin{algorithm}[H]
\caption{\textsc{ExecutableReflect}: the interventional phase of self-reflection}
\label{alg:evr}
\begin{algorithmic}[1]
\STATE sandbox $\leftarrow$ Python session preloaded with, for each case: its images, the gold answer, the model's wrong answer, and the code the model originally ran
\FOR{step $= 1, 2, \dots$ (bounded budget)}
   \STATE model writes investigation code \COMMENT{its own choice: re-run old code, print values and dtypes, render intermediates, test a fix against gold}
   \STATE execute; return the text output and the rendered images to the model
   \IF{model outputs skill edits}
      \RETURN edits \COMMENT{it decides on its own when the cause is verified}
   \ENDIF
\ENDFOR
\STATE \textbf{forced wrap-up:} one final call: ``based on what you verified, output the edits now''
\RETURN edits
\end{algorithmic}
\end{algorithm}

\end{document}